\documentclass{article}

\PassOptionsToPackage{numbers, compress}{natbib}

\usepackage[main,final]{neurips_2026}

\usepackage[utf8]{inputenc} 
\usepackage[T1]{fontenc}    
\usepackage{hyperref}       
\usepackage{url}            
\usepackage{booktabs}       
\usepackage{array}          
\usepackage{graphicx}       
\usepackage{pifont}
\usepackage{amsfonts}       
\usepackage{microtype}      
\usepackage{xcolor}

\usepackage{booktabs}
\usepackage{color}
\usepackage{xcolor}
\usepackage{amssymb}
\usepackage{amsmath}

\usepackage{amsmath}

\newcolumntype{L}[1]{>{\raggedright\arraybackslash}p{#1}}
\newcolumntype{C}[1]{>{\centering\arraybackslash}p{#1}}
\newcolumntype{R}[1]{>{\raggedleft\arraybackslash}p{#1}}

\definecolor{purple}{RGB}{128, 0, 128}

\newcommand{\est}[1]{\textcolor{black}{#1}}

\title{SliceWorld: A Predictive and Controllable World-State Model for CT Report Generation}

\author{Yuanhe Tian\textsuperscript{\ding{170}} \\
  \textsuperscript{\ding{170}}Zhongguancun Academy\\
  \textsuperscript{\ding{170}}\texttt{yhtian94@gmail.com} \\
  \And
  Yan Song\textsuperscript{\ding{171}}\thanks{Corresponding author.} \\
  \textsuperscript{\ding{171}}University of Science and Technology of China \\
  \textsuperscript{\ding{171}}\texttt{clksong@gmail.com} \\
}

\begin{document}

\maketitle

\begin{abstract}

CT report generation (CTRG) requires models to summarize three-dimensional anatomical context and pathological findings from hundreds of axial slices. Existing methods typically learn a direct image-to-text mapping, providing limited mechanisms for modeling how CT evidence evolves across slices or how reports respond to controlled changes in latent lesion-related factors. We propose SliceWorld, a CT-specific world-state framework that treats an axial CT scan as an ordered sequence along the z-axis. SliceWorld encodes prefix CT evidence into factor-aware latent states containing anatomy, lesion, and uncertainty components, and projects these states into world tokens used for multi-step future-slice feature prediction, lesion-factor intervention, and LLM-based report generation. The model is first pretrained on CT slice sequences with predictive, factor-aware, and counterfactual objectives, and is then fine-tuned on paired CT-report data. Experiments on M3D-Cap and CT-RATE show that SliceWorld improves natural language generation metrics and clinically oriented automatic evaluation. Further analyses demonstrate multi-horizon future-slice prediction, measurable factor alignment, reduced-slice robustness, and selective lesion-sensitive report modulation.\footnote{Code is available at \url{https://github.com/synlp/SliceWorld}.}

\end{abstract}

\section{Introduction}

Reports are the primary medium for communicating computed tomography (CT) findings and are central to diagnosis, treatment planning, and disease monitoring.
Automated reporting has therefore become an active research area, yet existing approaches often treat CTRG as direct image-to-text mapping and overlook the spatial and axial-sequential structure revealed by volumetric CT \cite{sloan2024automated, hamamci2024ct2rep, deng2024mvketr, li2025automatic, wang2025survey, kalisch2025ctgraph}.

Existing CT reporting approaches mainly follow two routes: global volume encoding for language-model generation \cite{hamamci2024ct2rep, chen2025large}, or sequential aggregation of axial slice features \cite{tian2025recurrent, tian2025feature}.
The former dilutes fine-grained anatomical localization, while the latter respects slice order but still treats the hidden state as a report-optimized feature vector rather than a predictive and intervenable CT state.
In parallel, the concept of \emph{world models} has emerged in reinforcement learning and video prediction as a powerful paradigm for building internal representations that capture the dynamics of an environment \cite{ha2018world, hafner2023worldmodels}. 
A world model learns not only to encode observations, but also to predict future observations, support intervention analysis, and provide a structured internal state for downstream decision making.
Translating this idea to medical imaging, specifically to CT report generation (CTRG), offers a compelling opportunity: we aim to construct an internal state that captures predictive slice-level visual evidence across adjacent axial slices and supports latent lesion-factor manipulation for report-level analysis.

\begin{figure}
    \centering
    \includegraphics[width=0.98\linewidth, trim=0 10 0 10]{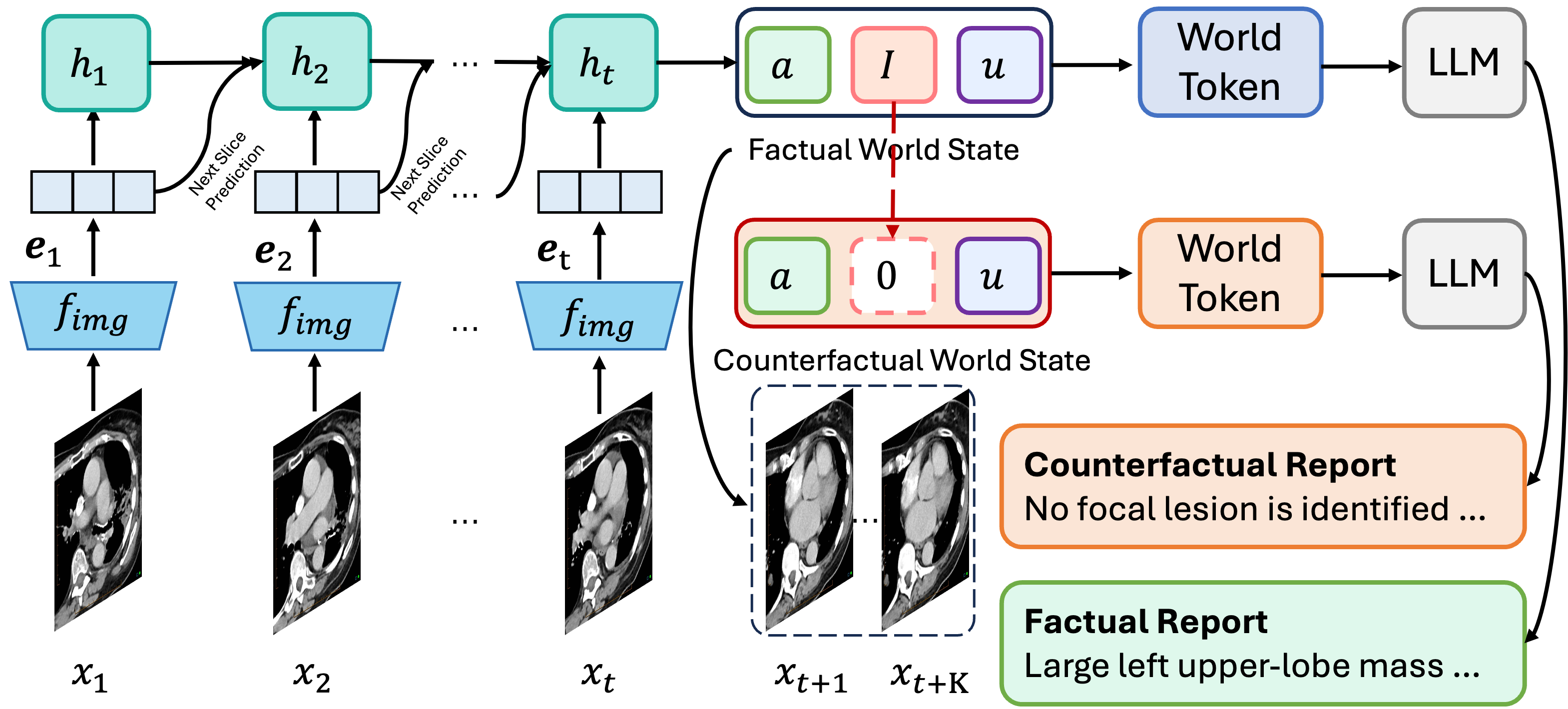}
    \caption{Overview of SliceWorld. Ordered axial CT slices are encoded into prefix states \(\mathbf{h}_t=f_{\mathrm{seq}}(\mathbf{e}_{1:t})\) without accessing future slices. Each \(\mathbf{h}_t\) is decomposed into anatomy \(\mathbf{a}_t\), lesion \(\mathbf{l}_t\), and uncertainty \(\mathbf{u}_t\) factors, projected into world tokens for LLM-based report generation, and optionally modified through lesion-factor neutralization to produce a latent lesion-neutral report variant.}
    \label{fig:model}
    \vspace{-0.3cm}
\end{figure}

In this paper, we propose \textbf{SliceWorld}, a framework that frames CTRG as a world-modeling problem over the axial slice sequence.
Rather than claiming a general-purpose physical simulator, SliceWorld studies a narrower but testable form of CT world modeling, where ordered axial evidence is compressed into a predictive and factor-aware state for report generation and controlled latent analysis.
Different from conventional sequential aggregators, SliceWorld learns a factor-aware state representation that encourages specialization of anatomy-, lesion-, and uncertainty-related signals.
The state evolution follows anatomical continuity priors and is trained with a suite of self- and weakly supervised world-model objectives, including multi-step future-slice feature prediction, factor-aware state learning, and counterfactual future-evidence prediction under lesion-factor intervention.
This design provides a structured interface between volumetric CT evidence and language generation, yielding three key capabilities: (i) diagnostic analysis, as the state components are probed for anatomy-, lesion-, and uncertainty-related signals; (ii) controllable analysis, where latent lesion-factor interventions produce lesion-suppressed report variants that are evaluated for non-target preservation; and (iii) partial-observation robustness, where the learned world state remains useful when only a subset of slices is available.
We evaluate SliceWorld not only with report-generation metrics, but also with future-slice feature prediction, factor-alignment probing, and lesion-factor intervention, each corresponding to a property implied by the proposed world-state formulation.
The results on benchmark CTRG datasets demonstrate the effectiveness of SliceWorld.

Our contributions are summarized as threefold:
\begin{itemize}
    \item We propose \textbf{SliceWorld}, a CT world model for CTRG, which learns a factor-aware state space over the axial slice sequence.
    \item We introduce a training framework that combines multi-step future-slice prediction, factor-aware state learning, and lesion-factor counterfactual prediction to encourage predictive and intervention-sensitive representations of axial CT structure.
    \item Extensive experiments demonstrate that SliceWorld achieves strong CTRG results, predicts future slice-level visual evidence beyond persistence baselines, shows measurable factor alignment, and supports lesion-sensitive latent report modulation.
\end{itemize}

\section{The Approach}

SliceWorld formulates CTRG as a world-modeling problem over ordered axial CT observations.\footnote{
In this CT-specific formulation, the environment is the patient-specific three-dimensional anatomical and pathological condition, and each axial slice provides a partial observation.
The latent state summarizes the observed axial prefix, predicts unobserved slice features, and supports lesion-factor intervention.
Prediction and intervention operate in feature and latent spaces, respectively.
}
Given a CT study $V = \{x_t\}_{t=1}^{T}$, where $x_t$ denotes the $t$-th axial slice and $T$ is the number of observed slices, SliceWorld first encodes the slice sequence into factor-aware world states.
At step $t$, the observation is the slice-level visual feature $\mathbf{e}_t$, and the world state $\mathbf{s}_t=[\mathbf{a}_t;\mathbf{l}_t;\mathbf{u}_t]$ summarizes the observed prefix, where $\mathbf{a}_t$, $\mathbf{l}_t$, and $\mathbf{u}_t$ denote the anatomy, lesion, and uncertainty factors, respectively.
SliceWorld then projects these states into a sequence of world tokens and generates the CT report from the projected tokens with an LLM.
Figure \ref{fig:model} illustrates this overall pipeline, including causal prefix encoding, factor-aware state construction, world-token projection, report generation, and lesion-factor intervention.
Formally, the overall generation process is expressed as
\begin{equation}
\widehat{R}
=
f_{\mathrm{gen}}
\big(
f_{\mathrm{tok}}
(
f_{\mathrm{wm}}(V)
)
\big)
\end{equation}
where $\widehat{R}$ denotes the generated CT report, $f_{\mathrm{wm}}$ denotes the world-state encoder that constructs factor-aware world states from the CT slice sequence, $f_{\mathrm{tok}}$ projects the world states into interface tokens, and $f_{\mathrm{gen}}$ denotes autoregressive report generation from the projected token sequence.
To strengthen SliceWorld before report-generation fine-tuning, we introduce a world-model pretraining stage on CT slice sequences.
This stage uses predictive and counterfactual objectives to learn representations that capture slice dynamics, semantic factors, and intervention-sensitive behavior.
The following subsections describe CT encoding and world-state construction, report generation from world tokens, and world-model pretraining with CTRG fine-tuning.

\subsection{CT Encoding and World-State Construction}

The world-state encoder $f_{\mathrm{wm}}$ aims to map the input CT slice sequence $\{x_t\}_{t=1}^{T}$ to a sequence of factor-aware world states.
Specifically, for each slice $x_t$, we apply an image encoder $f_{\mathrm{img}}$ to extract a slice-level visual representation by $\mathbf{e}_t = f_{\mathrm{img}}(x_t)$, where $\mathbf{e}_t$ denotes the visual feature of the $t$-th slice.
We then add a positional encoding $\mathbf{p}_t$ to preserve the relative axial location of each slice and feed the prefix sequence into a causal slice-sequence encoder by
\begin{equation}
\mathbf{h}_{t} = f_{\mathrm{seq}}(\mathbf{e}_{1:t} + \mathbf{p}_{1:t}) \quad t=1,\ldots,T
\end{equation}
where $\mathbf{h}_t \in \mathbb{R}^{d_h}$ denotes the contextualized representation at slice position $t$.
This causal construction ensures that $\mathbf{h}_t$ is computed only from $x_{1:t}$ and does not access future slices $x_{t+1:T}$ when it is used for multi-step future-slice prediction or downstream world-state construction.

Based on $\mathbf{h}_t$, SliceWorld constructs a factor-aware world state $\mathbf{s}_t=[\mathbf{a}_t;\mathbf{l}_t;\mathbf{u}_t]$ that contains anatomy state $\mathbf{a}_t$, lesion state $\mathbf{l}_t$, and uncertainty state $\mathbf{u}_t$ by
\begin{equation}
\mathbf{a}_t = f_{\mathrm{anat}}(\mathbf{h}_t) \quad
\mathbf{l}_t = f_{\mathrm{les}}(\mathbf{h}_t) \quad
\mathbf{u}_t = f_{\mathrm{unc}}(\mathbf{h}_t)
\end{equation}
To make lesion-related evidence explicit in the world model, we further attach a lesion-presence head to the lesion factor by
\begin{equation}
\widehat{m}_t = \sigma(f_{\mathrm{occ}}(\mathbf{l}_t))
\end{equation}
where \(\sigma(\cdot)\) is the sigmoid function and $\widehat{m}_t \in (0,1)$ denotes the predicted probability that slice $t$ is lesion-related.
This variable 
provides an explicit estimate of whether the current slice contains lesion-related evidence, which is later used in supervision and counterfactual learning.
Lesion annotations anchor the lesion factor, whereas the multi-step future-slice prediction objective uses feature targets without lesion labels.
The report generator receives the complete three-factor state, retaining anatomical context and prediction-difficulty information alongside lesion-related evidence.

\subsection{Report Generation from World Tokens}

After constructing the factor-aware world state $\mathbf{s}_t$ at each slice position, SliceWorld projects it into a world token $\mathbf{w}_t$ by
\begin{equation}
\mathbf{w}_t
=
f_{\mathrm{tok}}(\mathbf{s}_t)
\end{equation}
where $\mathbf{w}_t$ denotes the projected interface token used for language generation.\footnote{
The state $\mathbf{s}_t$ is the factor-aware world state that stores anatomy, lesion, and uncertainty information, while $\mathbf{w}_t$ is its tokenized representation passed to the language interface. 
Thus, $\mathbf{w}_t$ is not an additional recurrent state.
}
A two-layer multilayer perceptron $f_{\mathrm{lm}}$ further maps each world token into the hidden space of the LLM by $\mathbf{z}_t = f_{\mathrm{lm}}(\mathbf{w}_t)$, where $\mathbf{z}_t \in \mathbb{R}^{d_z}$ denotes the language-space token at slice position $t$.
The resulting token sequence $\mathbf{z}_{1:T}$ is fed into the LLM as visual-prefix context, and the report $\widehat{R}$ is generated through the standard autoregressive process.

The same state-to-token interface also supports lesion-factor intervention at inference time.
Given the factual world state $\mathbf{s}_t=[\mathbf{a}_t;\mathbf{l}_t;\mathbf{u}_t]$, SliceWorld constructs a latent lesion-neutral state by modifying only the lesion factor while keeping the anatomy and uncertainty factors unchanged by $\mathbf{s}_t^{\mathrm{lesion\text{-}zero}}
=
[\mathbf{a}_t;\mathbf{0};\mathbf{u}_t]$.
This latent state is then projected by the same token projector through $\mathbf{w}_t^{\mathrm{lesion\text{-}zero}}
=
f_{\mathrm{tok}}(\mathbf{s}_t^{\mathrm{lesion\text{-}zero}})$, where $\mathbf{w}_t^{\mathrm{lesion\text{-}zero}}$ denotes the world token produced after lesion-factor neutralization.
During factual and lesion-neutral generation, the decoder, prompt, decoding strategy, and input CT evidence are kept identical.
Thus, differences between the two generated reports are attributed to the latent lesion-factor modification.
Here, counterfactual analysis refers to the paired factual and lesion-factor-intervened latent states used to generate the two reports.

\subsection{World-Model Pretraining and Report Fine-Tuning}

SliceWorld is optimized with an overall objective composed of four terms: multi-step future-slice prediction (MFP), factor-aware state learning (FAS), counterfactual prediction (CF), and CTRG.
Therefore, the overall training loss $\mathcal{L}$ is written as
\begin{equation}
\mathcal{L}
=
\alpha_{\mathrm{mfp}}\mathcal{L}_{\mathrm{mfp}}
+
\alpha_{\mathrm{fas}}\mathcal{L}_{\mathrm{fas}}
+
\alpha_{\mathrm{cf}}\mathcal{L}_{\mathrm{cf}}
+
\alpha_{\mathrm{ctrg}}\mathcal{L}_{\mathrm{ctrg}}
\end{equation}
where $\mathcal{L}_{\mathrm{mfp}}$ denotes the multi-step feature prediction loss, $\mathcal{L}_{\mathrm{fas}}$ denotes the factor-aware state learning loss, $\mathcal{L}_{\mathrm{cf}}$ denotes the counterfactual prediction loss, and $\mathcal{L}_{\mathrm{ctrg}}$ denotes the CT report generation loss.
The coefficients $\alpha_{\mathrm{mfp}}$, $\alpha_{\mathrm{fas}}$, $\alpha_{\mathrm{cf}}$, and $\alpha_{\mathrm{ctrg}}$ control which objectives are active in each training stage.
The first three terms are used during world-model pretraining on CT slice sequences without text supervision, so that the factor-aware world states capture slice dynamics, lesion-related signals, and the intervention-sensitive effect of lesion factors on future observations.\footnote{In other words, $\alpha_{\mathrm{ctrg}}=0$ in the pretraining stage.}
The CTRG term $\mathcal{L}_{\mathrm{ctrg}}$ is then used during report fine-tuning to adapt the projected world-token interface to autoregressive report generation, where $\alpha_{\mathrm{mfp}}=\alpha_{\mathrm{fas}}=\alpha_{\mathrm{cf}}=0$ with the pretrained model parameters frozen, so the optimization focuses on aligning the pretrained world-token interface with paired report supervision rather than continuing to optimize auxiliary feature-space objectives.
The following texts present the four loss terms, where $\widehat{\mathbf{e}}_{t+k}^{(\cdot,k)}$ denotes a prediction of the future slice feature $\mathbf{e}_{t+k}$ with $k$ the future-prediction horizon, and the superscripts ($(h)$, $(w)$, and $(w,\mathrm{cf})$) specify whether the prediction is produced from the hidden state, the world token, or the counterfactual world token, respectively.

\paragraph{Multi-step Future-slice Prediction Loss.}

The MFP term trains SliceWorld to predict future slice-level visual features from the current prefix state.
It combines two prediction losses by
\begin{equation}
\mathcal{L}_{\mathrm{mfp}}
=
\lambda_h \mathcal{L}_{\mathrm{mfp}}^{h}
+
\lambda_w \mathcal{L}_{\mathrm{mfp}}^{w}
\end{equation}
where $\mathcal{L}_{\mathrm{mfp}}^{h}$ denotes the prediction loss from the sequence representation $\mathbf{h}_t$, $\mathcal{L}_{\mathrm{mfp}}^{w}$ denotes the prediction loss from the world token $\mathbf{w}_t$, and $\lambda_h$ and $\lambda_w$ are their corresponding loss weights.
A world model is expected to predict how the ordered observation sequence evolves after the current observation.
In SliceWorld, this property is enforced by predicting the next $K$ slice-level visual features from the current prefix state.
The first pathway predicts future slice features from the sequence representation:
\begin{equation}
\mathcal{L}_{\mathrm{mfp}}^{h}
=
\frac{1}{K(T-K)}
\sum_{t=1}^{T-K}
\sum_{k=1}^{K}
\left\|
\widehat{\mathbf{e}}_{t+k}^{(h,k)} - \mathbf{e}_{t+k}
\right\|_2^2
\end{equation}
where $\widehat{\mathbf{e}}_{t+k}^{(h,k)}=f_{\mathrm{pred},k}^{h}(\mathbf{h}_t)$ denotes the predicted feature of the $k$-step future slice from $\mathbf{h}_t$.
This objective trains the slice-sequence encoder to capture multi-step axial dynamics.
The second pathway predicts the same future slice features from the world token:
\begin{equation}
\mathcal{L}_{\mathrm{mfp}}^{w}
=
\frac{1}{K(T-K)}
\sum_{t=1}^{T-K}
\sum_{k=1}^{K}
\left\|
\widehat{\mathbf{e}}_{t+k}^{(w,k)} - \mathbf{e}_{t+k}
\right\|_2^2
\end{equation}
where $\widehat{\mathbf{e}}_{t+k}^{(w,k)}=f_{\mathrm{pred},k}^{w}(\mathbf{w}_t)$ denotes the predicted feature of the $k$-step future slice from $\mathbf{w}_t$.
This objective preserves future-prediction ability in the projected token interface, so that the sequence later used for report generation remains tied to the predictive world state.

\paragraph{Factor-aware State Loss.}

The FAS term encourages functional specialization of the anatomy, lesion, and uncertainty factors for stable anatomical context, lesion-related evidence, and prediction difficulty, respectively.
It is defined as
\begin{equation}
\mathcal{L}_{\mathrm{fas}}
=
\lambda_{\mathrm{smooth}}\mathcal{L}_{\mathrm{smooth}}
+
\lambda_{\mathrm{sparse}}\mathcal{L}_{\mathrm{sparse}}
+
\lambda_{\mathrm{unc}}\mathcal{L}_{\mathrm{unc}}
+
\lambda_{\mathrm{occ}}\mathcal{L}_{\mathrm{occ}}
\end{equation}
where $\mathcal{L}_{\mathrm{smooth}}$ regularizes anatomical continuity, $\mathcal{L}_{\mathrm{sparse}}$ regularizes lesion-factor sparsity, $\mathcal{L}_{\mathrm{unc}}$ aligns the uncertainty factor with future-prediction difficulty, and $\mathcal{L}_{\mathrm{occ}}$ supervises slice-level lesion presence.
The coefficients $\lambda_{\mathrm{smooth}}$, $\lambda_{\mathrm{sparse}}$, $\lambda_{\mathrm{unc}}$, and $\lambda_{\mathrm{occ}}$ are the corresponding loss weights.
The anatomical continuity term encourages the anatomy factor to vary smoothly across adjacent slices through
\begin{equation}
\mathcal{L}_{\mathrm{smooth}}
=
\frac{1}{T-1}
\sum_{t=2}^{T}
\left\|
\mathbf{a}_t - \mathbf{a}_{t-1}
\right\|_2^2
\label{eq:anatomy_smoothness}
\end{equation}
The lesion sparsity term encourages the lesion factor to remain sparse across the slice sequence through
\begin{equation}
\mathcal{L}_{\mathrm{sparse}}
=
\frac{1}{T}
\sum_{t=1}^{T}
\left\|
\mathbf{l}_t
\right\|_1
\label{eq:lesion_sparsity}
\end{equation}
The uncertainty-alignment term maps the uncertainty factor $\mathbf{u}_t$ to a scalar prediction-difficulty estimate and aligns it with the averaged future-prediction error through
\begin{equation}
\mathcal{L}_{\mathrm{unc}}
=
\frac{1}{T-K}
\sum_{t=1}^{T-K}
\left\|
g_{\mathrm{unc}}(\mathbf{u}_t)
-
\mathrm{sg}\!\left[
\frac{1}{K}
\sum_{k=1}^{K}
\mathrm{Err}
\big(
\widehat{\mathbf{e}}_{t+k}^{(h,k)},
\mathbf{e}_{t+k}
\big)
\right]
\right\|_2^2
\label{eq:uncertainty_alignment}
\end{equation}
where $g_{\mathrm{unc}}(\cdot)$ maps the uncertainty factor to a scalar, $\mathrm{Err}(\cdot,\cdot)$ denotes the normalized prediction error at each future horizon, and $\mathrm{sg}[\cdot]$ stops gradients through the averaged prediction-error target.
Finally, the lesion-presence term explicitly anchors the lesion factor to slice-level lesion-related evidence.
Let $m_t \in \{0,1\}$ denote the lesion-presence label for slice $t$.
We train the lesion-presence head with
\begin{equation}
\mathcal{L}_{\mathrm{occ}}
=
-\frac{1}{T}
\sum_{t=1}^{T}
\big(
m_t \log \widehat{m}_t
+
(1-m_t)\log(1-\widehat{m}_t)
\big)
\label{eq:lesion_presence}
\end{equation}
where $\widehat{m}_t$ is the lesion-presence probability predicted from the lesion factor $\mathbf{l}_t$.

\paragraph{Counterfactual Prediction Loss.}

The CF term trains SliceWorld to produce intervention-sensitive future predictions by comparing factual prediction with lesion-neutralized prediction.
It consists of an invariance loss on non-lesion slices and an effect loss on lesion-related slices:
\begin{equation}
\mathcal{L}_{\mathrm{cf}}
=
\lambda_{\mathrm{inv}}\mathcal{L}_{\mathrm{cf\text{-}inv}}
+
\lambda_{\mathrm{eff}}\mathcal{L}_{\mathrm{cf\text{-}eff}}
\end{equation}
where $\mathcal{L}_{\mathrm{cf\text{-}inv}}$ denotes the counterfactual invariance loss on non-lesion slices, $\mathcal{L}_{\mathrm{cf\text{-}eff}}$ denotes the counterfactual effect loss on lesion-related slices, and $\lambda_{\mathrm{inv}}$ and $\lambda_{\mathrm{eff}}$ are the corresponding loss weights.
A world model is expected not only to predict future observations, but also to respond consistently to interventions on its internal factors.
Given the factual world state $\mathbf{s}_t=[\mathbf{a}_t;\mathbf{l}_t;\mathbf{u}_t]$, SliceWorld constructs a counterfactual state by neutralizing only the lesion factor $\mathbf{s}_t^{\mathrm{cf}}
=
[\mathbf{a}_t; \mathbf{0}; \mathbf{u}_t]$,
where $\mathbf{s}_t^{\mathrm{cf}}$ denotes the lesion-neutralized counterfactual state.
This state is then projected into a counterfactual world token by $\mathbf{w}_t^{\mathrm{cf}}
=
f_{\mathrm{tok}}(\mathbf{s}_t^{\mathrm{cf}})$,
where $\mathbf{w}_t^{\mathrm{cf}}$ denotes the world token after lesion-factor neutralization.
The same multi-step prediction head is applied to the counterfactual token $\widehat{\mathbf{e}}_{t+k}^{(w,\mathrm{cf},k)}
=
f_{\mathrm{pred},k}^{w}(\mathbf{w}_t^{\mathrm{cf}})$,
where $\widehat{\mathbf{e}}_{t+k}^{(w,\mathrm{cf},k)}$ denotes the counterfactual prediction of the $k$-step future slice feature from the counterfactual world token.
For non-lesion slices, lesion-factor neutralization is expected to preserve the future prediction.
The invariance loss penalizes the discrepancy between factual and counterfactual predictions on non-lesion slices where $m_t=0$
\begin{equation}
\mathcal{L}_{\mathrm{cf\text{-}inv}}
=
\frac{
\sum_{t=1}^{T-K}
\sum_{k=1}^{K}
(1-m_t)
\left\|
\widehat{\mathbf{e}}_{t+k}^{(w,k)}
-
\widehat{\mathbf{e}}_{t+k}^{(w,\mathrm{cf},k)}
\right\|_2^2
}{
K\sum_{t=1}^{T-K}(1-m_t) + \epsilon
}
\label{eq:counterfactual_invariance}
\end{equation}
where $\epsilon$ is a small constant for numerical stability.
For lesion-related slices, lesion-factor neutralization is expected to reduce lesion-relevant predictive information and increase future-prediction error.
Since DeepLesion provides lesion-related slice ranges, $m_t$ indicates whether the current predictive state lies in a lesion-relevant interval, which reduces boundary mismatch when predicting future slice features $\mathbf{e}_{t+1:t+K}$.
We define the factual prediction error as
\begin{equation}
\ell_{t,k}^{\mathrm{fact}}
=
\left\|
\widehat{\mathbf{e}}_{t+k}^{(w,k)}
-
\mathbf{e}_{t+k}
\right\|_2^2
\label{eq:factual_prediction_error}
\end{equation}
and the counterfactual prediction error as
\begin{equation}
\ell_{t,k}^{\mathrm{cf}}
=
\left\|
\widehat{\mathbf{e}}_{t+k}^{(w,\mathrm{cf},k)}
-
\mathbf{e}_{t+k}
\right\|_2^2
\label{eq:counterfactual_prediction_error}
\end{equation}
where $\ell_{t,k}^{\mathrm{fact}}$ and $\ell_{t,k}^{\mathrm{cf}}$ denote the factual and counterfactual prediction errors for the $k$-step future slice, respectively.
We define the effect loss as a margin-based objective
\begin{equation}
\mathcal{L}_{\mathrm{cf\text{-}eff}}
=
\frac{
\sum_{t=1}^{T-K}
\sum_{k=1}^{K}
m_t
\max\big(0, \delta + \ell_{t,k}^{\mathrm{fact}} - \ell_{t,k}^{\mathrm{cf}}\big)
}{
K\sum_{t=1}^{T-K} m_t + \epsilon
}
\label{eq:counterfactual_effect}
\end{equation}
where $\delta$ is the margin that requires the counterfactual prediction error to be larger than the factual prediction error on lesion-related slices.
This objective encourages the lesion factor to act as an intervention-sensitive component of future prediction rather than as a passive feature channel.

\paragraph{CTRG Fine-tuning Loss.}

The CTRG term is the standard autoregressive negative log-likelihood over the target report $R=(r_1,\ldots,r_N)$.
This term optimizes the language decoder to generate the target report from the world-token sequence.

\section{Experiment Settings}

\subsection{Dataset}

We use three datasets with different roles in the overall pipeline.
DeepLesion \cite{yan2018deeplesion} is used for world-model pretraining, while M3D-Cap \cite{bai2024m3d} and CT-RATE \cite{hamamci2026ctrate} are used for CTRG.
These datasets provide CT data in different formats.
DeepLesion provides lesion-centric axial CT slices with lesion annotations and ordered slice context.
M3D-Cap provides CT image-text pairs across multiple anatomical regions and CT views.
CT-RATE provides chest CT volumes in HU-valued NIfTI format, paired with radiology reports and abnormality labels.
Since SliceWorld takes an ordered image sequence as input, we convert each dataset into this unified representation.
For DeepLesion, each CT series is sorted by axial order and used as a slice sequence for world-model pretraining.
For M3D-Cap, we follow the official image-text construction and use the ordered CT image group associated with each annotated sample when multiple images are available; samples with a single image are treated as length-one sequences.
For CT-RATE, we convert each HU-valued NIfTI volume into a sequence of windowed RGB slice images before feeding it to the visual encoder.
DeepLesion is used only for world-model pretraining because it does not provide paired full radiology reports.
Its lesion annotations and slice ranges support learning slice dynamics, factor-aware states, slice-level lesion presence, and lesion-factor counterfactual prediction.
M3D-Cap evaluates adaptation to heterogeneous CT image-text supervision, while CT-RATE evaluates volume-level chest CT report generation.
For all datasets, we follow the official split protocols.
The CT-RATE HU-to-image preprocessing protocol is described in Appendix \ref{app:ctrate_preprocessing}.
DeepLesion series construction and slice-level lesion-label construction are described in Appendix \ref{app:deeplesion_labels}.
Additional details for the report datasets, including M3D-Cap and CT-RATE split usage, are reported in Appendix \ref{app:report_dataset_details}.

\begin{table}[t]
\centering
\caption{
Results of existing studies and our approach with different backbone LLMs on M3D-Cap.
``$^\dag$'' marks scores reported in Tian et al. \cite{tian2025feature}, and models without any marker use their original reported scores.
LLM size reports the language backbone size when applicable.
}
\small
\setlength{\tabcolsep}{2.5pt}
\begin{tabular}{l | r | ccccc}
\toprule
 & LLM size & BLEU-1 & ROUGE-1 & METEOR & BERT-score & RaTEScore (F1) \\
\midrule
MedVInT-TD$^{\dag}$ \cite{zhang2023pmc} 
& 7B & 11.7 & 16.2 & 11.4 & 87.7 & -- \\
Med-Flamingo$^{\dag}$ \cite{moor2023med} 
& 9B & 12.4 & 16.8 & 11.7 & 88.0 & -- \\
RadFM$^{\dag}$ \cite{wu2023towards}
& 14B & 13.1 & 17.4 & 13.0  & 88.3 & -- \\
M3D \cite{bai2024m3d}
& 7B & 15.2 & 19.6 & 14.4 & 88.5 & -- \\
CT2Rep$^{\dag}$ \cite{hamamci2024ct2rep}
& -- & 16.2 & 18.8 & 14.1 & 88.5 & -- \\
S-LMR \cite{tian2025feature}
& 2B & 17.4 & 19.3 & 14.8 & 88.8 & 36.0 \\
\midrule
Ours (Qwen3-1.7B)
& 2B & 18.1 & 20.5 & 15.3 & 89.3 & 36.7 \\
Ours (Qwen3-4B)
& 4B & \textbf{18.5} & \textbf{20.8} & \textbf{15.6} & \textbf{89.6} & \textbf{37.1} \\
Ours (Ministral-3-3B)
& 3B & 18.3 & 20.7 & 15.4 & 89.4 & 36.9 \\
\bottomrule
\end{tabular}
\label{tab:sota}
\vspace{-0.3cm}
\end{table}

\subsection{Baselines}

We compare SliceWorld against strong baselines from recent CT and medical vision-language generation studies.
The compared approaches include general medical multimodal models, such as MedVInT-TD~\cite{zhang2023pmc}, Med-Flamingo~\cite{moor2023med}, and RadFM~\cite{wu2023towards}, as well as CT-specific report generation approaches, including M3D~\cite{bai2024m3d}, CT2Rep~\cite{hamamci2024ct2rep}, CT2-CHAT~\cite{hamamci2024developing}, and S-LMR~\cite{tian2025feature}.
These baselines cover global volume-to-text modeling, slice-sequential report generation, multimodal foundation models, and anatomy-lesion decomposition-based report generation.

\subsection{Implementation Details}

SliceWorld uses DINOv2 ViT \cite{oquab2023dinov2} as the slice-level visual encoder and a randomly initialized Mamba sequence encoder \cite{gu2023mamba} for axial sequence modeling.
A cross-attention compression module aggregates DINOv2 patch tokens into one representation per observed slice.
The Mamba encoder is implemented as a unidirectional causal scan.
For language decoding, the world-token sequence is mapped to the decoder hidden space by a two-layer multilayer perceptron and used as a continuous prefix to the textual input.
We use Qwen3-1.7B, Qwen3-4B \cite{yang2025qwen3}, and Ministral-3-3B \cite{liu2026ministral} as decoder backbones.
During report fine-tuning, we freeze the pretrained world-state encoder, apply LoRA \cite{hu2022lora} to the LLMs, train the projection layers that connect world tokens to the decoder, and use greedy decoding at inference time.
All model hyperparameters, LLM backbone details, loss weights, and training settings are reported in Appendix \ref{app:impl_details}, with computational cost summarized in Appendix \ref{app:compute}.
All experiments are conducted on NVIDIA A40 GPUs.

We evaluate model performance with two metric categories: standard natural language generation metrics and clinically oriented metrics.
The first measures surface-form and semantic similarity, while the second measures whether clinically relevant findings are preserved.
For M3D-Cap, we report BLEU-1, ROUGE-1, METEOR, BERTScore, and RaTEScore (F1); for CT-RATE, we report BLEU-1--4, ROUGE-L, METEOR, GREEN, and RaTEScore (F1).
In addition, we also perform human evaluation, where we randomly sample 100 cases from the test set and ask two experts to assign a score ranging from 1 to 5 (higher is better) to the report based on its quality, and report the average score.
The detailed evaluation protocol, metric definitions, and rationale for dataset-specific metric choices are provided in Appendix \ref{app:eval_details}.

\begin{table*}[t]
\centering
\caption{
Results of existing studies and our approach with different backbone LLMs on CT-RATE.
``$^\dag$'' and ``$^\ddag$'' mark scores reported in Tian et al. \cite{tian2025feature} and Hamamci et al. \cite{hamamci2026ctrate}, respectively.
Models without any marker use their original reported scores.
LLM size reports the language backbone size when applicable.
}
\small
\setlength{\tabcolsep}{2.2pt}
\begin{tabular}{l | r | cccccc || c | c}
\toprule
& LLM size & BL-1  & BL-2  & BL-3  & BL-4 & R-L & MT  & Green & RaTEScore (F1) \\
\midrule
MedVInT-TD$^{\dag}$ \cite{zhang2023pmc} 
& 7B & 33.9 & 26.7 & 21.8 & 18.3 & 30.2 & 18.4 & 40.2 & 12.1 \\
Med-Flamingo$^{\dag}$ \cite{moor2023med} 
& 9B & 34.9 & 27.4 & 22.5 & 19.0 & 31.1 & 19.1 & 42.4 & 13.8 \\
RadFM$^{\dag}$ \cite{wu2023towards}
& 14B & 36.4 & 28.0 & 22.9 & 19.3 & 31.7 & 19.9 & 43.3 & 15.4 \\
CT2Rep$^{\ddag}$ \cite{hamamci2024ct2rep}
& -- & 37.2 & 29.2 & 24.4 & 21.3 & 36.2 & 19.7 & 48.7 & 16.0 \\
CT2-CHAT$^{\ddag}$ \cite{hamamci2024developing}
& 8B & 37.3 & 28.4 & 23.1 & 19.8 & 32.6 & 21.5 & 43.6 & 18.4 \\
S-LMR$^{\dag}$ \cite{tian2025feature}
& 2B & 38.1 & 29.6 & 24.9 & 21.7 & 37.0 & 21.9 & 49.4 & 18.9 \\
\midrule
Ours (Qwen3-1.7B)
& 2B & {38.7} & {30.4} & {25.4} & {22.2} & {37.7} & {22.4} & {50.1} & {20.1} \\
Ours (Qwen3-4B)
& 4B & \textbf{39.4} & \textbf{{30.9}} & \textbf{{26.0}} & \textbf{22.6} & \textbf{{38.2}} & \textbf{{23.0}} & \textbf{50.7} & \textbf{20.6}\\
Ours (Ministral-3-3B)
& 3B & 39.0 & 30.6 & 25.7 & 22.4 & 38.0 & 22.7 & 50.4 & 20.2 \\
\bottomrule
\end{tabular}
\label{tab:sota-2}
\vspace{-0.3cm}
\end{table*}

\section{Results and Analyses}

\subsection{Overall Results}

The CTRG results of our approach and existing representative models on M3D-Cap and CT-RATE are in Table \ref{tab:sota} and Table \ref{tab:sota-2}, respectively.
We make two observations.
First, SliceWorld outperforms the compared representative CTRG approaches on both datasets.
Most prior approaches mainly treat CTRG as direct image-to-text generation or slice-feature aggregation, without explicitly learning a predictive and intervention-sensitive world state.
The improvements over these baselines therefore suggest that world-modeling provides an effective representation interface for CTRG.
Notably, SliceWorld also surpasses several baselines with substantially larger LLM backbones, such as Med-Flamingo (9B), RadFM (14B), and CT2-CHAT (8B), indicating that the gain is not merely a result of scaling the decoder.
Second, SliceWorld obtains strong results with different backbone LLMs.
Across M3D-Cap and CT-RATE, these variants remain competitive or superior on both standard NLG metrics and clinically oriented automatic metrics.

\subsection{Ablation Study}

To examine the contribution of each component in SliceWorld, we conduct an ablation study of our approach with Qwen3-1.7B on CT-RATE\footnote{We also report the results on M3D-Cap under the same setting in Appendix Table \ref{tab:m3d_same_backbone_appendix}.} with different model variants under the same settings.
\emph{Direct CTRG} removes the DeepLesion world-model pretraining stage and fine-tunes the same SliceWorld architecture end-to-end on the CTRG dataset, with all model components updated rather than randomly frozen.
Starting from Direct CTRG, we then add different pretraining objectives on DeepLesion to form the remaining variants.
All pretrained variants use the same DeepLesion series, preprocessing, and pretraining budget, and keep the pretrained visual/world-state components frozen during report fine-tuning.
\emph{+Recon}\footnote{To separate architectural gains from additional pretraining data, the reconstruction and predictive variants use the same DeepLesion budget and differ only in whether the auxiliary objective reconstructs current-slice evidence or predicts future axial evidence.} adds a data-matched reconstruction-pretraining stage that uses an MAE-style current-slice reconstruction target instead of MFP.
\emph{+MFP} replaces this reconstruction target with multi-step future-slice prediction while retaining the same three-factor architecture and disabling factor-aware objectives, lesion-specific supervision, and counterfactual learning.
\emph{+FAS} builds on +MFP by further adding anatomy smoothness, lesion sparsity, uncertainty calibration, and lesion-presence supervision.
\emph{Full SliceWorld} builds on +FAS by additionally adding factual/lesion-zero counterfactual future prediction.
Table \ref{tab:ablation_ctrate} shows a consistent step-by-step improvement.
Compared with Direct CTRG, +Recon tests whether gains are explained by the additional DeepLesion data alone.
+Recon improves over Direct CTRG, but +MFP performs better with the same DeepLesion data and number of pretraining epochs, indicating a contribution from predictive learning beyond reconstruction.
Because +MFP uses no lesion-specific labels, the comparison also shows that part of the reporting gain arises without lesion supervision.
Under the same DeepLesion pretraining data, +FAS further improves performance, suggesting that factor-aware learning improves clinically relevant consistency, and Full SliceWorld performs best after adding CF, showing that counterfactual learning complements the predictive and factor-aware objectives.
This monotonic trend supports that each component of the proposed world-model training contributes to final CTRG performance, which is further analyzed through prediction, factor-alignment, and intervention studies below.

\begin{table*}[t]
\centering
\caption{
Ablation study of our approach on CT-RATE with Qwen3-1.7B. All settings share the same architecture and report-generation training setup; +Recon, +MFP, +FAS, and Full SliceWorld use the same DeepLesion data. ``\(^{*}\)'' marks paired bootstrap significance over Direct CTRG (\(p\le0.05\)).
}
\small
\setlength{\tabcolsep}{2.8pt}
\begin{tabular}{l | cccccc || c | c || c}
\toprule
& BL-1 & BL-2 & BL-3 & BL-4 & R-L & MT & Green & RaTEScore (F1) & Expert (1--5) \\
\midrule
Direct CTRG & {36.9} & {28.5} & {23.4} & {20.1} & {35.4} & {20.8} & {47.2} & {17.5} & 3.3 \\
\quad +Recon & {37.3} & {29.0} & {23.9} & {20.6} & {36.0} & {21.2} & {47.8} & {18.0} & 3.3 \\
\quad +MFP & {37.8} & {29.4} & {24.4} & {21.4} & {36.5} & {21.5} & {48.5} & {18.5} & 3.5 \\
\quad +FAS & {38.2} & {30.1} & {25.1} & {22.0} & {37.2} & {22.0} & {49.4} & {19.3} & 3.5 \\
SliceWorld 
& \textbf{{38.7}}$^{*}$ & \textbf{{30.4}}$^{*}$ & \textbf{{25.4}}$^{*}$ & \textbf{{22.2}}$^{*}$ & \textbf{{37.7}}$^{*}$ & \textbf{{22.4}}$^{*}$ & \textbf{{50.1}}$^{*}$ & \textbf{{20.1}}$^{*}$ & \textbf{3.7} \\
\bottomrule
\end{tabular}
\label{tab:ablation_ctrate}
\vspace{-0.3cm}
\end{table*}

\subsection{Effect of CT World Modeling}

\begin{figure*}[t]
\centering
\includegraphics[width=0.98\textwidth]{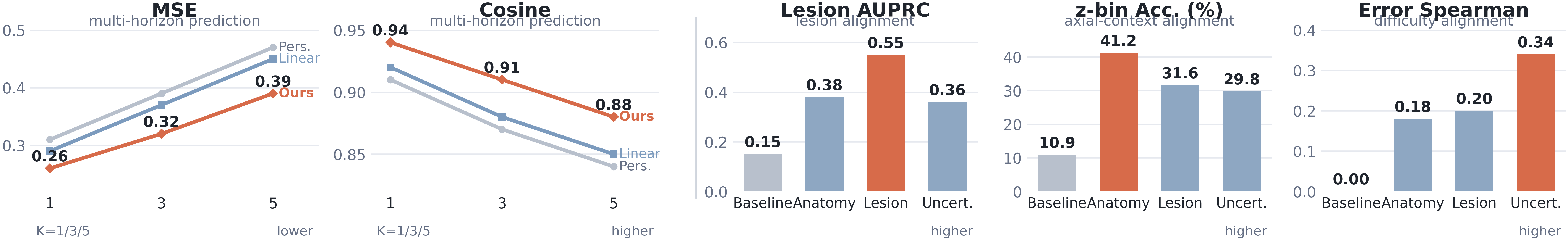}
\caption{State-level diagnostics of SliceWorld with Qwen3-1.7B. The left two panels show multi-horizon prefix-only future-slice feature prediction on CT-RATE for \(K\in\{1,3,5\}\). The right three panels show factor-alignment diagnostics on held-out DeepLesion sequences by comparing each state factor with simple statistical baselines. Higher is better except for MSE.}
\label{fig:state_diagnostics}
\vspace{-0.2cm}
\end{figure*}

One of the most essential characteristics of a world model is its capability to predict future observations, which correspond to upcoming slices in the CT volume in our setting.
Therefore, we explore whether our approach preserves multi-step future-slice prediction ability.
That is, for each position $t$, the model receives only the prefix slices $x_{1:t}$ and predicts the visual features of future slices \(x_{t+1:t+K}\).
Since no future slice is encoded, good performance across multiple future horizons indicates that the world state carries predictive spatial information rather than only serving the CTRG decoder.
Considering that adjacent CT slices are naturally similar, we compare SliceWorld (using the Qwen3-1.7B backbone) with two CT-continuity baselines, i.e., the persistence baseline and the linear extrapolation baseline.
The persistence baseline directly uses the current slice feature for every future horizon, i.e., \(\widehat{\mathbf{e}}_{t+k}=\mathbf{e}_t\).
The linear extrapolation baseline assumes that the recent feature displacement continues linearly, i.e., \(\widehat{\mathbf{e}}_{t+k}=\mathbf{e}_t+k(\mathbf{e}_t-\mathbf{e}_{t-1})\).
We report mean squared error (MSE), where lower is better, and cosine similarity, where higher is better, between the predicted and ground-truth visual features.
Figure \ref{fig:state_diagnostics} visualizes the multi-horizon prediction trend in the left two panels.
SliceWorld achieves lower MSE and higher cosine similarity than both baselines, which suggests that the final CTRG model retains a predictive representation of axial CT structure, rather than learning only a direct mapping from CT features to report text.
These comparisons assess prediction beyond simple axial-continuity rules, while Table \ref{tab:ablation_ctrate} examines how the training objectives affect downstream report generation.

\subsection{Effect of Factor-Aware State Learning}

We test whether factor-aware learning assigns clinically relevant information to the intended factors rather than merely improving CTRG scores.
Keeping SliceWorld fixed, we extract \(\mathbf{a}_t\), \(\mathbf{l}_t\), and \(\mathbf{u}_t\) on held-out DeepLesion sequences and train the same shallow probe family for each factor.
The lesion probe predicts slice-level lesion labels $m_t$ and is evaluated by the area under the precision-recall curve (AUPRC), while the axial-context probe predicts ten normalized $z$-bins and is evaluated by accuracy.
The uncertainty probe estimates the averaged future-prediction difficulty $d_t=\frac{1}{K}\sum_{k=1}^{K}\|\widehat{\mathbf{e}}_{t+k}^{(w,k)}-\mathbf{e}_{t+k}\|_2^2$ and is evaluated by Spearman correlation.
The right three panels of Figure \ref{fig:state_diagnostics} show a consistent diagonal trend above simple statistical baselines: \(\mathbf{l}_t\) is best for lesion AUPRC, \(\mathbf{a}_t\) for \(z\)-bin accuracy, and \(\mathbf{u}_t\) for error Spearman.
All three factors are projected from the same causal hidden state $\mathbf{h}_t$, while the factor-aware objectives encourage distinct functional roles.
The nonzero off-diagonal scores indicate residual information sharing among the factors, so the probes support preferential functional specialization rather than statistical independence. Appendix Figure \ref{fig:qualitative_appendix} provides qualitative examples.

\subsection{Selective Lesion Intervention}

We further test whether manipulating the lesion factor produces selective clinical report changes.
For each CT-RATE study, we compare a factual report from $\mathbf{w}_{1:T}$ with lesion-zero and uncertainty-zero variants, where \(\mathbf{s}_t^{\mathrm{lesion\text{-}zero}}=[\mathbf{a}_t;\mathbf{0};\mathbf{u}_t]\) and \(\mathbf{s}_t^{\mathrm{unc\text{-}zero}}=[\mathbf{a}_t;\mathbf{l}_t;\mathbf{0}]\).
We measure target change and mention removal on a narrow focal-lesion target set, and preservation/hallucination on non-target findings.\footnote{We interpret lesion-zero generation as latent report modulation rather than physical lesion removal, and therefore evaluate it through target change, non-target preservation, and negative-control perturbations rather than as clinical counterfactual ground truth.}
Table \ref{tab:lesion_zero_main} shows that lesion-zero moderately but selectively reduces focal-lesion content in lesion-positive cases, while behaving approximately as a no-op in lesion-negative cases and preserving most non-target findings.
The uncertainty-zero control produces much smaller target changes, indicating that the effect is tied to the lesion factor rather than arbitrary latent perturbation.
The negative controls test factor selectivity, while non-target preservation and hallucination rates measure the breadth of the induced report changes.
To further evaluate whether lesion-factor modulation introduces unsafe report changes, Appendix Table \ref{tab:intervention_validity_proxy} reports automatic clinical-validity proxies based on CT-RATE labels and RadBERT-CT extraction.
These proxies show that lesion-zero keeps the overall clinical-label agreement nearly unchanged, has low false-negative risk on non-target findings, and rarely flips non-target labels, supporting that the intervention changes focal-lesion content without broadly corrupting the report.
Appendix Figure \ref{fig:cf_examples} gives illustrative examples.

\begin{table}[t]
\centering
\caption{Selective lesion-factor intervention on CT-RATE. Target metrics use a narrow focal-lesion target set; preservation and hallucination are computed on non-target findings. Uncertainty-zero is included as a negative control.}
\small
\vspace{-0.1cm}
\setlength{\tabcolsep}{2pt}
\begin{tabular}{l | cccc}
\toprule
Group & Target chg. (\%) & Mention rem. (\%) & Preserv. (\%) & Halluc. (\%) \\
\midrule
Positive, lesion-zero & \est{58.4} & \est{52.7} & \est{94.8} & \est{2.2} \\
Positive, uncertainty-zero & \est{9.6} & \est{6.8} & \est{96.7} & \est{1.4} \\
Negative, lesion-zero & \est{4.1} & \est{3.0} & \textbf{\est{97.3}} & \textbf{\est{1.2}} \\
Negative, uncertainty-zero & \est{2.9} & \est{2.2} & \textbf{\est{97.9}} & \textbf{\est{1.1}} \\
\bottomrule
\end{tabular}
\label{tab:lesion_zero_main}
\vspace{-0.2cm}
\end{table}

\section{Related Work}

Radiology report generation has been studied extensively in chest X-ray and is increasingly explored in CT, where recent approaches improve image-text alignment through volume-to-text modeling, slice-sequential encoding, anatomy-guided structures, multimodal foundation models, or selective state-space backbones~\cite{chen2020generating, chen2021cross, liu2023systematic, sloan2024automated, liu2024bootstrapping, sun2024r2genmamba, deng2024mvketr, bai2024m3d, tian2024diffusion, hamamci2024ct2rep, chen2025large, claessens2025spectre, tian2025feature, tian2025extractive, tian2025recurrent, kalisch2025ctgraph, li2025automatic,wang2025survey, blankemeier2026merlin}.
These studies improve report quality by better encoding volumetric context or by separating anatomical and pathological cues, but they still mainly learn a direct mapping from CT representations to text rather than an explicit predictive world representation.
More broadly, general vision-language studies model relations among image-text samples~\cite{fu2023learning} and predict structured concepts for image captioning~\cite{wang2023improving}.
In parallel, world models have been developed for simulation, prediction, and counterfactual reasoning in sequential environments~\cite{ha2018world, hafner2023worldmodels}, while their use in medical imaging remains limited and has mostly focused on representation learning rather than report generation~\cite{yue2025chexworld}.
SliceWorld connects these lines of studies by learning a lesion-aware world representation over axial CT slices that is predictive of future observations and usable for latent lesion-factor report modulation.

\section{Conclusion}

We presented SliceWorld, a CT world model for report generation that treats an axial CT scan as an ordered observation sequence and learns a predictive, factor-aware latent state over slice-level visual evidence.
By combining multi-step future-slice feature prediction, factor-aware state learning, lesion-factor counterfactual prediction, and an LLM world-token interface, SliceWorld moves CTRG beyond direct volume-to-text mapping toward a predictive and controllable representation interface.
Experiments on M3D-Cap and CT-RATE show consistent improvements over representative CTRG baselines across standard NLG metrics and clinically oriented evaluations.
Further ablations and analyses show that the learned state retains prefix-only future-slice prediction ability, exhibits measurable factor alignment, improves robustness under partial slice observation, and supports selective lesion-factor report modulation.
These results suggest that CT-specific world-state modeling provides a useful framework for improving volumetric report generation and for analyzing how latent lesion-related evidence affects generated reports.

\section*{Acknowledgements}

This work is supported by the Key Research and Development Program of Anhui Province, China (Grant No. 202423k09020030) and Zhongguancun Academy (Grant No. XTS0043).

\bibliographystyle{unsrtnat}
\bibliography{checked}

@article{yan2018deeplesion,
  title={DeepLesion: automated mining of large-scale lesion annotations and universal lesion detection with deep learning},
  author={Yan, Ke and Wang, Xiaosong and Lu, Le and Summers, Ronald M},
  journal={Journal of medical imaging},
  volume={5},
  number={3},
  pages={036501--036501},
  year={2018},
}

@article{ha2018world,
  title={World models},
  author={Ha, David and Schmidhuber, J{\"u}rgen},
  journal={arXiv preprint arXiv:1803.10122},
  volume={2},
  number={3},
  pages={440},
  year={2018}
}

@article{hafner2023worldmodels,
  title={Mastering diverse domains through world models},
  author={Hafner, Danijar and Pasukonis, Jurgis and Ba, Jimmy and Lillicrap, Timothy},
  journal={arXiv preprint arXiv:2301.04104},
  year={2023}
}

@inproceedings{hu2022lora,
  title={Lora: Low-rank adaptation of large language models.},
  author={Hu, Edward J and Shen, Yelong and Wallis, Phillip and Allen-Zhu, Zeyuan and Li, Yuanzhi and Wang, Shean and Wang, Liang and Chen, Weizhu and others},
  journal={Iclr},
  volume={1},
  number={2},
  pages={3},
  year={2022}
}

@article{deng2024mvketr,
  title={Mvketr: chest CT report generation with multi-view perception and knowledge enhancement},
  author={Deng, Xiwei and He, Xianchun and Bao, Jianfeng and Zhou, Yudan and Cai, Shuhui and Cai, Congbo and Chen, Zhong},
  journal={IEEE Journal of Biomedical and Health Informatics},
  year={2025},
}

@inproceedings{kalisch2025ctgraph,
  title={CT-GRAPH: Hierarchical Graph Attention Network for Anatomy-Guided CT Report Generation},
  author={Kalisch, Hamza and H{\"o}rst, Fabian and Kleesiek, Jens and Herrmann, Ken and Seibold, Constantin},
  booktitle={Proceedings of the IEEE/CVF International Conference on Computer Vision},
  pages={6775--6784},
  year={2025}
}

@article{tian2025recurrent,
  title={Recurrent Visual Feature Extraction and Stereo Attentions for CT Report Generation},
  author={Tian, Yuanhe and Mao, Lei and Song, Yan},
  journal={arXiv preprint arXiv:2506.19665},
  year={2025}
}

@article{blankemeier2026merlin,
  title={Merlin: a computed tomography vision--language foundation model and dataset},
  author={Blankemeier, Louis and Kumar, Ashwin and Cohen, Joseph Paul and Liu, Jiaming and Liu, Longchao and Van Veen, Dave and Gardezi, Syed Jamal Safdar and Yu, Hongkun and Paschali, Magdalini and Chen, Zhihong and others},
  journal={Nature},
  pages={1--11},
  year={2026},
}

@article{hamamci2026ctrate,
  title={Generalist foundation models from a multimodal dataset for 3D computed tomography},
  author={Hamamci, Ibrahim Ethem and Er, Sezgin and Wang, Chenyu and Almas, Furkan and Simsek, Ayse Gulnihan and Esirgun, Sevval Nil and Dogan, Irem and Durugol, Omer Faruk and Hou, Benjamin and Shit, Suprosanna and others},
  journal={Nature Biomedical Engineering},
  pages={1--19},
  year={2026},
}

@inproceedings{yue2025chexworld,
  title={Chexworld: Exploring image world modeling for radiograph representation learning},
  author={Yue, Yang and Wang, Yulin and Tao, Chenxin and Liu, Pan and Song, Shiji and Huang, Gao},
  booktitle={Proceedings of the Computer Vision and Pattern Recognition Conference},
  pages={20778--20788},
  year={2025}
}

@article{claessens2025spectre,
  title={Scaling Self-Supervised and Cross-Modal Pretraining for Volumetric CT Transformers},
  author={Claessens, Cris and Viviers, Christiaan and D'Amicantonio, Giacomo and Bondarev, Egor and van der Sommen, Fons},
  journal={arXiv preprint arXiv:2511.17209},
  year={2025}
}

@article{sun2024r2genmamba,
  title={R2gen-mamba: A selective state space model for radiology report generation},
  author={Sun, Yongheng and Lee, Yueh Z and Woodard, Genevieve A and Zhu, Hongtu and Lian, Chunfeng and Liu, Mingxia},
  booktitle={2025 IEEE 22nd International Symposium on Biomedical Imaging (ISBI)},
  pages={1--4},
  year={2025},
}

@article{zhang2023pmc,
  title={Pmc-vqa: Visual instruction tuning for medical visual question answering},
  author={Zhang, Xiaoman and Wu, Chaoyi and Zhao, Ziheng and Lin, Weixiong and Zhang, Ya and Wang, Yanfeng and Xie, Weidi},
  journal={arXiv preprint arXiv:2305.10415},
  year={2023}
}

@article{gu2023mamba,
  title={Mamba: Linear-time sequence modeling with selective state spaces},
  author={Gu, Albert and Dao, Tri},
  journal={arXiv preprint arXiv:2312.00752},
  year={2023}
}

@article{zhang2019bertscore,
  title={Bertscore: Evaluating text generation with bert},
  author={Zhang, Tianyi and Kishore, Varsha and Wu, Felix and Weinberger, Kilian Q and Artzi, Yoav},
  journal={arXiv preprint arXiv:1904.09675},
  year={2019}
}

@inproceedings{zhao2024ratescore,
  title={RaTEScore: A Metric for Radiology Report Generation},
  author={Zhao, Weike and Wu, Chaoyi and Zhang, Xiaoman and Zhang, Ya and Wang, Yanfeng and Xie, Weidi},
  booktitle={Proceedings of the 2024 Conference on Empirical Methods in Natural Language Processing},
  pages={15004--15019},
  year={2024}
}

@article{yang2025qwen3,
  title={Qwen3 technical report},
  author={Yang, An and Li, Anfeng and Yang, Baosong and Zhang, Beichen and Hui, Binyuan and Zheng, Bo and Yu, Bowen and Gao, Chang and Huang, Chengen and Lv, Chenxu and others},
  journal={arXiv preprint arXiv:2505.09388},
  year={2025}
}

@article{liu2026ministral,
  title={Ministral 3},
  author={Liu, Alexander H and Khandelwal, Kartik and Subramanian, Sandeep and Jouault, Victor and Rastogi, Abhinav and Sad{\'e}, Adrien and Jeffares, Alan and Jiang, Albert and Cahill, Alexandre and Gavaudan, Alexandre and others},
  journal={arXiv preprint arXiv:2601.08584},
  year={2026}
}

@article{ostmeier2024green,
  title={Green: Generative radiology report evaluation and error notation},
  author={Ostmeier, Sophie and Xu, Justin and Chen, Zhihong and Varma, Maya and Blankemeier, Louis and Bluethgen, Christian and Michalson, Arne Edward and Moseley, Michael and Langlotz, Curtis and Chaudhari, Akshay S and others},
  journal={arXiv preprint arXiv:2405.03595},
  year={2024}
}

@inproceedings{lin-2004-rouge,
    title = "{{ROUGE}: A Package for Automatic Evaluation of Summaries}",
    author = "Lin, Chin-Yew",
    booktitle = "Text Summarization Branches Out",
    month = jul,
    year = "2004",
    address = "Barcelona, Spain",
    pages = "74--81"
}

@inproceedings{banerjee2005meteor,
  title={METEOR: An automatic metric for MT evaluation with improved correlation with human judgments},
  author={Banerjee, Satanjeev and Lavie, Alon},
  booktitle={Proceedings of the acl workshop on intrinsic and extrinsic evaluation measures for machine translation and/or summarization},
  pages={65--72},
  year={2005}
}

@inproceedings{papineni2002bleu,
  title={{BLEU: a method for automatic evaluation of machine translation}},
  author={Papineni, Kishore and Roukos, Salim and Ward, Todd and Zhu, Wei-Jing},
  booktitle={Proceedings of the 40th annual meeting of the Association for Computational Linguistics},
  pages={311--318},
  year={2002}
}

@article{oquab2023dinov2,
  title={Dinov2: Learning robust visual features without supervision},
  author={Oquab, Maxime and Darcet, Timoth{\'e}e and Moutakanni, Th{\'e}o and Vo, Huy and Szafraniec, Marc and Khalidov, Vasil and Fernandez, Pierre and Haziza, Daniel and Massa, Francisco and El-Nouby, Alaaeldin and others},
  journal={arXiv preprint arXiv:2304.07193},
  year={2023}
}

@inproceedings{moor2023med,
  title={Med-flamingo: a multimodal medical few-shot learner},
  author={Moor, Michael and Huang, Qian and Wu, Shirley and Yasunaga, Michihiro and Dalmia, Yash and Leskovec, Jure and Zakka, Cyril and Reis, Eduardo Pontes and Rajpurkar, Pranav},
  booktitle={Machine Learning for Health (ML4H)},
  pages={353--367},
  year={2023}
}

@article{wu2023towards,
  title={Towards generalist foundation model for radiology},
  author={Wu, Chaoyi and Zhang, Xiaoman and Zhang, Ya and Wang, Yanfeng and Xie, Weidi},
  journal={arXiv preprint arXiv:2308.02463},
  year={2023}
}

@article{hamamci2024developing,
  title={Developing Generalist Foundation Models from a Multimodal Dataset for 3D Computed Tomography},
  author={Hamamci, Ibrahim Ethem and Er, Sezgin and Almas, Furkan and Simsek, Ayse Gulnihan and Esirgun, Sevval Nil and Dogan, Irem and Dasdelen, Muhammed Furkan and Durugol, Omer Faruk and Wittmann, Bastian and Amiranashvili, Tamaz and others},
  year={2024}
}

@article{bai2024m3d,
  title={{M3D: Advancing 3d medical image analysis with multi-modal large language models}},
  author={Bai, Fan and Du, Yuxin and Huang, Tiejun and Meng, Max Q-H and Zhao, Bo},
  journal={arXiv preprint arXiv:2404.00578},
  year={2024}
}

@article{li2025automatic,
  title={Automatic radiology report generation with deep learning: a comprehensive review of methods and advances},
  author={Li, Yilin and Kong, Chao and Zhao, Guosheng and Zhao, Zijian},
  journal={Artificial Intelligence Review},
  volume={58},
  number={11},
  pages={344},
  year={2025},
}

@article{sloan2024automated,
  title={Automated radiology report generation: A review of recent advances},
  author={Sloan, Phillip and Clatworthy, Philip and Simpson, Edwin and Mirmehdi, Majid},
  journal={IEEE Reviews in Biomedical Engineering},
  volume={18},
  pages={368--387},
  year={2024},
  publisher={IEEE}
}

@article{tian2025feature,
  title={Feature decomposition via shared low-rank matrix recovery for ct report generation},
  author={Tian, Yuanhe and Song, Yan},
  journal={IEEE Transactions on Medical Imaging},
  year={2025}
}

@article{wang2025survey,
  title={A survey of deep-learning-based radiology report generation using multimodal inputs},
  author={Wang, Xinyi and Figueredo, Grazziela and Li, Ruizhe and Zhang, Wei Emma and Chen, Weitong and Chen, Xin},
  journal={Medical Image Analysis},
  volume={103},
  pages={103627},
  year={2025},
}

@inproceedings{hamamci2024ct2rep,
  title={{CT2Rep}: Automated radiology report generation for 3d medical imaging},
  author={Hamamci, Ibrahim Ethem and Er, Sezgin and Menze, Bjoern},
  booktitle={International Conference on Medical Image Computing and Computer-Assisted Intervention},
  pages={476--486},
  year={2024},
}

@article{chen2025large,
  title={Large language model with region-guided referring and grounding for CT report generation},
  author={Chen, Zhixuan and Bie, Yequan and Jin, Haibo and Chen, Hao},
  journal={IEEE Transactions on Medical Imaging},
  year={2025},
}

@article{tian2025extractive,
  title={Extractive radiology reporting with memory-based cross-modal representations},
  author={Tian, Yuanhe and Yan, Zexuan and Lyu, Nenan and Song, Yan},
  journal={IEEE Transactions on Medical Imaging},
  year={2025}
}

@inproceedings{fu2023learning,
  title={Learning semantic relationship among instances for image-text matching},
  author={Fu, Zheren and Mao, Zhendong and Song, Yan and Zhang, Yongdong},
  booktitle={2023 IEEE/CVF Conference on Computer Vision and Pattern Recognition (CVPR)},
  pages={15159--15168},
  year={2023}
}

@inproceedings{wang2023improving,
  title={Improving image captioning via predicting structured concepts},
  author={Wang, Ting and Chen, Weidong and Tian, Yuanhe and Song, Yan and Mao, Zhendong},
  booktitle={Proceedings of the 2023 conference on empirical methods in natural language processing},
  pages={360--370},
  year={2023}
}

@inproceedings{tian2024diffusion,
  title={Diffusion networks with task-specific noise control for radiology report generation},
  author={Tian, Yuanhe and Xia, Fei and Song, Yan},
  booktitle={Proceedings of the 32nd ACM International Conference on Multimedia},
  pages={1771--1780},
  year={2024}
}

@inproceedings{liu2024bootstrapping,
  title={Bootstrapping large language models for radiology report generation},
  author={Liu, Chang and Tian, Yuanhe and Chen, Weidong and Song, Yan and Zhang, Yongdong},
  booktitle={Proceedings of the AAAI conference on artificial intelligence},
  volume={38},
  number={17},
  pages={18635--18643},
  year={2024}
}

@article{liu2023systematic,
  title={A systematic review of deep learning-based research on radiology report generation},
  author={Liu, Chang and Tian, Yuanhe and Song, Yan},
  journal={arXiv preprint arXiv:2311.14199},
  year={2023}
}

@inproceedings{chen2021cross,
  title={Cross-modal memory networks for radiology report generation},
  author={Chen, Zhihong and Shen, Yaling and Song, Yan and Wan, Xiang},
  booktitle={Proceedings of the 59th annual meeting of the association for computational linguistics and the 11th international joint conference on natural language processing (volume 1: long papers)},
  pages={5904--5914},
  year={2021}
}

@inproceedings{chen2020generating,
  title={Generating radiology reports via memory-driven transformer},
  author={Chen, Zhihong and Song, Yan and Chang, Tsung-Hui and Wan, Xiang},
  booktitle={Proceedings of the 2020 conference on empirical methods in natural language processing (EMNLP)},
  pages={1439--1449},
  year={2020}
}

\newpage
\appendix

\section{CT-RATE HU-to-Image Preprocessing}
\label{app:ctrate_preprocessing}

CT-RATE provides 3D NIfTI volumes whose voxel values are already stored in Hounsfield units (HU) as int16 values.
We therefore read the NIfTI array directly as HU and do not reapply the DICOM rescale intercept or slope.
For each volume, we identify the axial dimension, iterate over all slices, and convert each slice into a three-channel uint8 image using fixed CT windows:
lung window \((\mathrm{WL}=-600,\mathrm{WW}=1500)\) as the red channel, soft-tissue window \((\mathrm{WL}=40,\mathrm{WW}=400)\) as the green channel, and bone window \((\mathrm{WL}=400,\mathrm{WW}=1500)\) as the blue channel.
For each window, HU values are clipped to \([\mathrm{WL}-\mathrm{WW}/2,\mathrm{WL}+\mathrm{WW}/2]\), linearly scaled to \([0,255]\), and cast to uint8.
Each slice is resized to \(224\times224\) using area interpolation, and all slices are saved as an array of shape \((N,224,224,3)\).

\section{DeepLesion Pretraining Data Construction}
\label{app:deeplesion_labels}

DeepLesion contains 32,120 axial CT slices from 10,594 studies of 4,427 patients, with 32,735 lesion annotations in total.
In addition to lesion bounding boxes and size measurements, it provides patient-, study-, and series-level metadata, key slices, and lesion-related slice ranges.
For DeepLesion pretraining, each CT series is sorted by axial slice order and treated as one observation sequence.
We follow the official patient-level split and organize each \((\mathrm{Patient}, \mathrm{Study}, \mathrm{Series})\) tuple as one ordered axial slice sequence.
We assign the slice-level lesion label \(m_t=1\) when slice \(t\) contains an annotated lesion box or falls within the lesion-related slice range of an annotated lesion in the same series.
Slices outside annotated lesion ranges are used as weak negatives with \(m_t=0\).
Because DeepLesion is lesion-centric and absence of annotation does not guarantee absence of pathology, we use these labels only for world-model pretraining and factor-alignment diagnostics, and we describe the resulting factors as aligned rather than perfectly disentangled.

\section{Report Dataset Details}
\label{app:report_dataset_details}

M3D-Cap is a large-scale CT image-text dataset with 120,092 image-text pairs and official train/validation/test splits of 116,092/2,000/2,000.
Its samples cover multiple anatomical regions and CT views, providing diverse text supervision for adapting the learned world representation to report generation.
Unlike CT-RATE, M3D-Cap is not organized as full-volume radiology reports for every sample.
We therefore follow its official image-text benchmark protocol: when a sample provides multiple ordered CT images from the same annotated case/view, we treat them as an ordered slice group; when only one image is available, we use it as a length-one sequence.
This keeps the evaluation consistent with prior M3D-Cap studies while allowing SliceWorld to use the same ordered-sequence interface.
CT-RATE is a chest CT-report dataset that pairs thoracic CT volumes with radiology reports and abnormality labels.
We follow the released split protocol for CT-RATE and use the corresponding training, validation, and test partitions for model development and evaluation.
The CT-RATE HU-to-image conversion is described separately in Appendix \ref{app:ctrate_preprocessing}.

\section{Additional Implementation Details}
\label{app:impl_details}

\paragraph{Backbone details.}
The DINOv2 ViT slice encoder has 24 Transformer blocks, hidden dimension 1,024, and 16 attention heads.
The Mamba sequence encoder has 12 layers with hidden dimension \(d_{\mathrm{model}}=512\), state dimension \(d_{\mathrm{state}}=16\), convolution width \(d_{\mathrm{conv}}=4\), and expansion factor 2.
The state-factor dimensions are \(d_{\mathrm{anat}}=256\), \(d_{\mathrm{lesion}}=192\), and \(d_{\mathrm{unc}}=64\), and the world-token dimension is 512.
The decoder backbones are Qwen3-1.7B, Qwen3-4B, and Ministral-3-3B.
Qwen3-1.7B has 28 Transformer layers with hidden size 2,048, 16 attention heads, and 8 key-value heads; Qwen3-4B has 36 layers with hidden size 2,560, 32 attention heads, and 8 key-value heads; Ministral-3-3B has 26 layers with hidden size 3,072, 32 attention heads, and 8 key-value heads.

\paragraph{Slice-token compression.}
For each CT slice \(x_t\), DINOv2 outputs patch tokens \(\mathbf{P}_t \in \mathbb{R}^{M\times d_v}\).
We use a learnable slice query \(\mathbf{q}_{\mathrm{slc}}\) to attend to these patch tokens through a multi-head cross-attention module:
\begin{equation}
\widetilde{\mathbf{e}}_t
=
\mathrm{CrossAttn}(\mathbf{q}_{\mathrm{slc}}, \mathbf{P}_t, \mathbf{P}_t)
\quad
\mathbf{e}_t = W_e \mathrm{LN}(\widetilde{\mathbf{e}}_t)
\end{equation}
This produces one slice-level feature \(\mathbf{e}_t\) per observed axial slice before sequence modeling.

\paragraph{LLM prefix construction.}
After factor-aware states are projected into world tokens, we map them to the LLM hidden dimension as \(\mathbf{z}_{1:T}=f_{\mathrm{lm}}(\mathbf{w}_{1:T})\).
During report fine-tuning, these continuous embeddings are prepended to the textual prompt embeddings and then followed by target report tokens:
\begin{equation}
\mathbf{E}_{\mathrm{in}}
=
[\mathbf{z}_{1:T};\mathbf{E}_{\mathrm{prompt}};\mathbf{E}_{\mathrm{report}}]
\end{equation}
The autoregressive loss is applied only to report tokens, while the world-token prefix and prompt tokens serve as conditioning context.

\paragraph{Training hyperparameters.}
World-model pretraining uses AdamW with learning rate \(1\times 10^{-4}\), weight decay 0.01, batch size 8, dropout 0.1, maximum sequence length 256, maximum sampled slices 196, and 3 epochs.
For MFP, the future-prediction horizon is set to \(K=5\), i.e., each valid prefix state predicts the next five slice-level DINOv2 features.
Report fine-tuning uses AdamW with learning rate \(2\times 10^{-5}\), weight decay 0.01, LoRA rank 16, LoRA \(\alpha=32\), and LoRA dropout 0.05.
For M3D-Cap, we use batch size 4, maximum sampled slices 196, and maximum text length 256.
For CT-RATE, we use batch size 1 with gradient accumulation 2, maximum sampled slices 480, and maximum text length 384.

\paragraph{Loss weights.}
During world-model pretraining, the active top-level objectives use \(\alpha_{\mathrm{mfp}}=\alpha_{\mathrm{fas}}=\alpha_{\mathrm{cf}}=1\).
During report fine-tuning, the active report-generation objective uses \(\alpha_{\mathrm{ctrg}}=1\).
All component weights inside the active objectives are set to 1; the \(\alpha\) coefficients indicate whether an objective is active in the current optimization phase.
For the MFP term, \(\lambda_h=\lambda_w=1\).
For the FAS term, \(\lambda_{\mathrm{smooth}}=\lambda_{\mathrm{sparse}}=\lambda_{\mathrm{unc}}=\lambda_{\mathrm{occ}}=1\), and the lesion-presence BCE uses unit positive-class weight.
For the CF term, \(\lambda_{\mathrm{inv}}=\lambda_{\mathrm{eff}}=1\), and the hinge margin is \(\delta=0.1\).

\section{Computational Cost}
\label{app:compute}

All reported runs use NVIDIA A40 GPUs.
Table \ref{tab:compute_ablation} summarizes the cost of the CT-RATE ablation settings in Table \ref{tab:ablation_ctrate}.
The main additional cost of SliceWorld comes from the world-model pretraining stage, while report fine-tuning cost remains similar across ablation variants because the same DINOv2 encoder, Mamba encoder, Qwen3-1.7B decoder, LoRA configuration, and CT-RATE split are used.

\begin{table}[h]
\centering
\caption{Computational cost of the CT-RATE ablation settings with Qwen3-1.7B on A40 GPUs.}
\small
\setlength{\tabcolsep}{5pt}
\begin{tabular}{l | cccc}
\toprule
Training setting & Pretrain GPU h & Fine-tune GPU h & Total GPU h & Peak memory (GB) \\
\midrule
Direct CTRG & {0.0} & {38.5} & {38.5} & {36.2} \\
+Recon & {20.8} & {38.7} & {59.5} & {36.9} \\
+MFP & {21.4} & {38.7} & {60.1} & {37.0} \\
+FAS & {24.9} & {38.9} & {63.8} & {37.5} \\
Full SliceWorld & {28.6} & {39.0} & {67.6} & {38.1} \\
\bottomrule
\end{tabular}
\label{tab:compute_ablation}
\end{table}

Table \ref{tab:compute_backbone} reports the computational cost across the three LLM backbones.
Since the visual encoder and world-state encoder are shared, larger language backbones mainly increase LoRA fine-tuning memory and decoding latency.
The corresponding report-generation performance is already reported in Table \ref{tab:sota} and Table \ref{tab:sota-2}.

\begin{table*}[h]
\centering
\caption{Backbone-level computational cost. Memory and time are measured on A40 GPUs under the CT-RATE fine-tuning and inference setting.}
\small
\setlength{\tabcolsep}{5pt}
\begin{tabular}{l | cccc}
\toprule
Backbone & Params & Fine-tune GPU h & Peak memory (GB) & Sec./study \\
\midrule
Qwen3-1.7B & {1.7B} & {39.0} & {38.1} & {2.4} \\
Qwen3-4B & {4.0B} & {62.5} & {57.8} & {4.1} \\
Ministral-3-3B & {3.3B} & {54.2} & {51.6} & {3.5} \\
\bottomrule
\end{tabular}
\label{tab:compute_backbone}
\end{table*}

\section{Additional Evaluation Details}
\label{app:eval_details}

\paragraph{Report-generation metrics.}
We use standard natural language generation metrics, including BLEU~\cite{papineni2002bleu}, ROUGE~\cite{lin-2004-rouge}, METEOR~\cite{banerjee2005meteor}, and BERTScore~\cite{zhang2019bertscore}.
BLEU and ROUGE mainly measure lexical overlap, METEOR further accounts for stemming and synonym matching, and BERTScore measures semantic similarity with contextualized token embeddings.
These metrics evaluate surface-form and semantic agreement, but they do not fully capture clinical correctness.

\paragraph{Clinically oriented metrics.}
For M3D-Cap, we additionally report RaTEScore~\cite{zhao2024ratescore}, following the common evaluation practice on this benchmark.
For CT-RATE, we report GREEN~\cite{ostmeier2024green}, which is designed for clinically oriented automatic evaluation in chest imaging, and RaTEScore (F1) for finding-level report consistency.
We compute abnormality-level precision, recall, and F1 for CT-RATE.
We do not apply GREEN to M3D-Cap because GREEN is developed for chest imaging, whereas M3D-Cap contains CT data from multiple anatomical regions.

\paragraph{Expert evaluation protocol.}
To complement automatic clinical metrics, we perform a blinded expert evaluation protocol on 100 randomly sampled CT-RATE studies.
Two clinical experts independently score each generated report on a 1--5 clinical correctness scale, where 1 indicates mostly incorrect or unsafe output and 5 indicates clinically accurate and complete output.
The system identity is hidden and report order is randomized.

\paragraph{Significance testing.}
For results marked with \(^{*}\) in the ablation study, we use paired bootstrap resampling over the test studies with 10,000 resamples and compare Full SliceWorld against Direct CTRG under the same evaluation pipeline.
We mark significance for the metrics when the one-sided bootstrap \(p\)-value is at most 0.05.
We do not use \(^{*}\) markers in the SOTA comparison tables because several prior scores are taken from existing studies and do not provide paired per-study outputs under our pipeline.

\section{Additional Experiment Results}
\label{app:additional_ctrg}

We include additional experiment results that complement the main analyses, including backbone-controlled CTRG checks, partial-observation robustness, and sensitivity analyses for the world-model training design.

\paragraph{Same-backbone M3D-Cap ablation.}
Table \ref{tab:m3d_same_backbone_appendix} compares direct report fine-tuning with progressive SliceWorld training variants on M3D-Cap under the same Qwen3-1.7B decoder, DINOv2 encoder, Mamba sequence encoder, LoRA configuration, split, and decoding setting.
Direct CTRG removes the DeepLesion world-model pretraining stage and updates all model components end-to-end on M3D-Cap, while +MFP, +FAS, and Full SliceWorld use the same DeepLesion series, preprocessing, and pretraining budget before freezing the pretrained visual/world-state components during report fine-tuning.
The comparison shows that world-model pretraining improves CTRG beyond the shared visual and language backbone.

\begin{table}[h]
\centering
\caption{Same-backbone CTRG ablation on M3D-Cap with Qwen3-1.7B.}
\small
\setlength{\tabcolsep}{3pt}
\begin{tabular}{l | ccccc}
\toprule
Training setting & BLEU-1 & ROUGE-1 & METEOR & BERTScore & RaTEScore (F1) \\
\midrule
Direct CTRG & {16.9} & {19.3} & {14.5} & {88.7} & {35.5} \\
+MFP & \est{17.5} & \est{19.9} & \est{14.9} & \est{89.0} & \est{36.1} \\
+FAS & \est{17.8} & \est{20.2} & \est{15.1} & \est{89.1} & \est{36.4} \\
Full SliceWorld & \textbf{{18.1}} & \textbf{{20.5}} & \textbf{{15.3}} & \textbf{{89.3}} & \textbf{{36.7}} \\
\bottomrule
\end{tabular}
\label{tab:m3d_same_backbone_appendix}
\end{table}

\paragraph{M3D-Cap single-image and multi-slice groups.}
Because M3D-Cap contains both length-one image-text samples and samples with multiple ordered CT images, we further split its test set into single-image and multi-slice input groups.
This analysis tests whether the axial world-state interface remains useful when multiple ordered observations are available, while respecting that M3D-Cap is not a full-volume report dataset.
As shown in Table \ref{tab:m3dcap_slice_groups}, SliceWorld improves over Direct CTRG in both groups, with a slightly larger gain in the multi-slice group where axial evidence is available.
Since the single-image group is small, we treat this as a supporting analysis rather than the main evidence.

\begin{table}[h]
\centering
\caption{M3D-Cap single-image versus multi-slice input-group analysis with Qwen3-1.7B.}
\small
\setlength{\tabcolsep}{3pt}
\begin{tabular}{l | l | r | cccc}
\toprule
Model & Group & N & BLEU-1 & ROUGE-1 & METEOR & RaTEScore \\
\midrule
Direct CTRG & single-image & \est{65} & \est{10.2} & \est{15.1} & \est{13.0} & \est{31.2} \\
Full SliceWorld & single-image & \est{65} & \est{10.8} & \est{15.8} & \est{13.6} & \est{32.0} \\
Direct CTRG & multi-slice & \est{1935} & \est{17.1} & \est{19.4} & \est{14.6} & \est{35.6} \\
Full SliceWorld & multi-slice & \est{1935} & \est{18.3} & \est{20.7} & \est{15.4} & \est{36.9} \\
\bottomrule
\end{tabular}
\label{tab:m3dcap_slice_groups}
\end{table}

\paragraph{Reduced-slice CT-RATE inference.}
Table \ref{tab:partial_slice_appendix} evaluates CT-RATE report generation when only a uniformly sampled subset of slices is provided at inference time.
Both models use the same Qwen3-1.7B decoder and the same decoding protocol.
SliceWorld degrades more slowly as the slice budget decreases, supporting the robustness of the learned world state under partial axial observation.

\begin{table}[h]
\centering
\caption{Reduced-slice CTRG robustness on CT-RATE with Qwen3-1.7B.}
\small
\setlength{\tabcolsep}{5pt}
\begin{tabular}{l | c | ccc}
\toprule
Model & Slice budget & BL-1 & GREEN & RaTEScore (F1) \\
\midrule
Direct CTRG & 100\% & {36.9} & {47.2} & {17.5} \\
Direct CTRG & 50\% & {35.8} & {46.1} & {16.4} \\
Direct CTRG & 25\% & {34.6} & {44.8} & {15.3} \\
Full SliceWorld & 100\% & \textbf{{38.7}} & \textbf{{50.1}} & \textbf{{20.1}} \\
Full SliceWorld & 50\% & \textbf{{37.8}} & \textbf{{49.0}} & \textbf{{19.2}} \\
Full SliceWorld & 25\% & \textbf{{36.6}} & \textbf{{47.6}} & \textbf{{18.1}} \\
\bottomrule
\end{tabular}
\label{tab:partial_slice_appendix}
\end{table}

\paragraph{Loss-weight sensitivity.}
We evaluate CT-RATE sensitivity to the top-level FAS and CF weights while keeping the main experiment setting fixed.
We keep \(\alpha_{\mathrm{mfp}}=1\) because multi-step future-slice prediction is the core predictive objective, and keep \(\alpha_{\mathrm{ctrg}}=1\) during report fine-tuning.
Table \ref{tab:alpha_sensitivity} sweeps \(\alpha_{\mathrm{fas}},\alpha_{\mathrm{cf}}\in\{0.5,0.8,1.0\}\) using the same CT-RATE setup as Table \ref{tab:ablation_ctrate}.
The results improve as either FAS or CF receives larger weight, and the default \((\alpha_{\mathrm{fas}},\alpha_{\mathrm{cf}})=(1.0,1.0)\) achieves the strongest result, supporting the usefulness of both losses.

\begin{table}[h]
\centering
\caption{CT-RATE sensitivity to FAS and CF loss weights with Qwen3-1.7B. \(\alpha_{\mathrm{mfp}}\) and \(\alpha_{\mathrm{ctrg}}\) are fixed to 1 in their active training stages.}
\small
\setlength{\tabcolsep}{5pt}
\begin{tabular}{cc | ccc}
\toprule
\(\alpha_{\mathrm{fas}}\) & \(\alpha_{\mathrm{cf}}\) & BL-1 & GREEN & RaTEScore (F1) \\
\midrule
\est{0.5} & \est{0.5} & \est{37.9} & \est{48.9} & \est{18.9} \\
\est{0.5} & \est{0.8} & \est{38.1} & \est{49.2} & \est{19.2} \\
\est{0.5} & \est{1.0} & \est{38.2} & \est{49.4} & \est{19.3} \\
\est{0.8} & \est{0.5} & \est{38.2} & \est{49.3} & \est{19.2} \\
\est{0.8} & \est{0.8} & \est{38.4} & \est{49.6} & \est{19.6} \\
\est{0.8} & \est{1.0} & \est{38.5} & \est{49.8} & \est{19.8} \\
\est{1.0} & \est{0.5} & \est{38.4} & \est{49.7} & \est{19.6} \\
\est{1.0} & \est{0.8} & \est{38.6} & \est{49.9} & \est{19.9} \\
\est{1.0} & \est{1.0} & \textbf{\est{38.7}} & \textbf{\est{50.1}} & \textbf{\est{20.1}} \\
\bottomrule
\end{tabular}
\label{tab:alpha_sensitivity}
\end{table}

\begin{figure}[h]
\centering
\includegraphics[width=0.92\linewidth, trim=0 50 0 0]{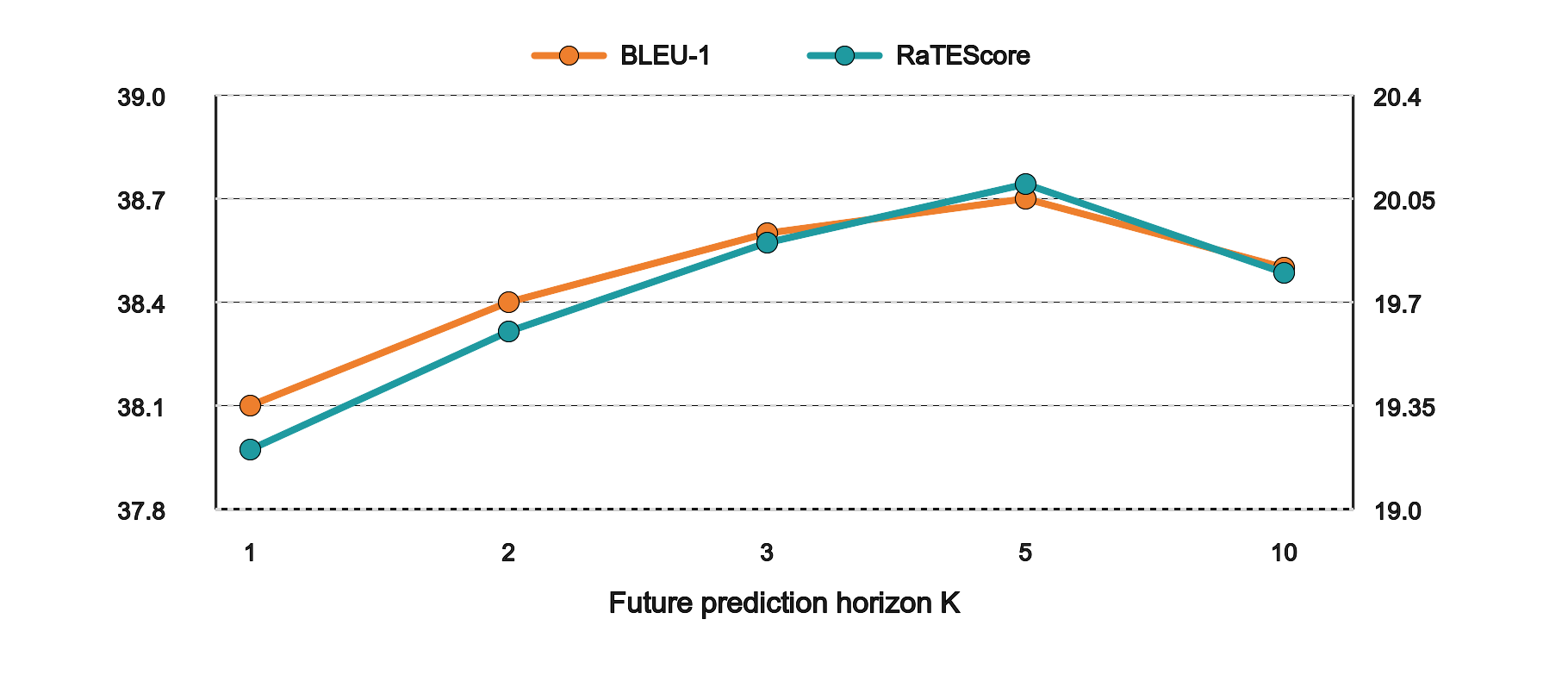}
\caption{CT-RATE sensitivity to the future-prediction horizon \(K\). We report BLEU-1 and RaTEScore with Qwen3-1.7B. The default \(K=5\) gives the strongest balance between report-generation quality and clinically oriented correctness.}
\label{fig:nsp_horizon_sensitivity}
\end{figure}

\paragraph{Future-horizon sensitivity.}
We also evaluate how the multi-step future-prediction horizon affects downstream CT-RATE report generation.
Figure \ref{fig:nsp_horizon_sensitivity} reports the result for \(K\in\{1,2,3,5,10\}\) using one standard NLG metric and one clinically oriented metric.
Short horizons provide a useful predictive signal, but \(K=5\) performs best because it encourages the state to capture a slightly broader axial context without forcing predictions too far away from the observed prefix.
The mild decrease at \(K=10\) suggests that very long feature-space prediction can introduce unnecessary difficulty while still remaining competitive with shorter horizons.

\paragraph{Intervention-validity proxies.}

Table \ref{tab:intervention_validity_proxy} complements the main selective lesion-intervention results with automatic clinical-validity proxies.
We compare factual and intervention reports using RadBERT-CT labels and CT-RATE gold labels, and compute false-negative risk and contradiction on non-target findings.
These auxiliary results check whether lesion-factor modulation preserves non-target clinical content rather than merely changing focal-lesion words.

\begin{table}[h]
\centering
\caption{Automatic intervention-validity proxies on CT-RATE. Clinical F1 \(\Delta\) compares intervention reports against factual reports using RadBERT-CT labels and CT-RATE gold labels; false-negative risk and contradiction are computed on non-target findings.}
\small
\setlength{\tabcolsep}{4pt}
\begin{tabular}{l | cccc}
\toprule
Intervention & Clinical F1 \(\Delta\) & Non-target F1 \(\Delta\) & FN risk (\%) & Contradiction (\%) \\
\midrule
Lesion-zero & \est{-0.002} & \est{-0.001} & \est{1.8} & \est{4.9} \\
Uncertainty-zero & \est{-0.001} & \est{-0.001} & \est{1.2} & \est{3.7} \\
\bottomrule
\end{tabular}
\label{tab:intervention_validity_proxy}
\end{table}

\section{Additional Qualitative Cases}
\label{app:qualitative_cases}

\begin{figure*}[!h]
\centering
\includegraphics[width=\linewidth, trim=50 30 50 30]{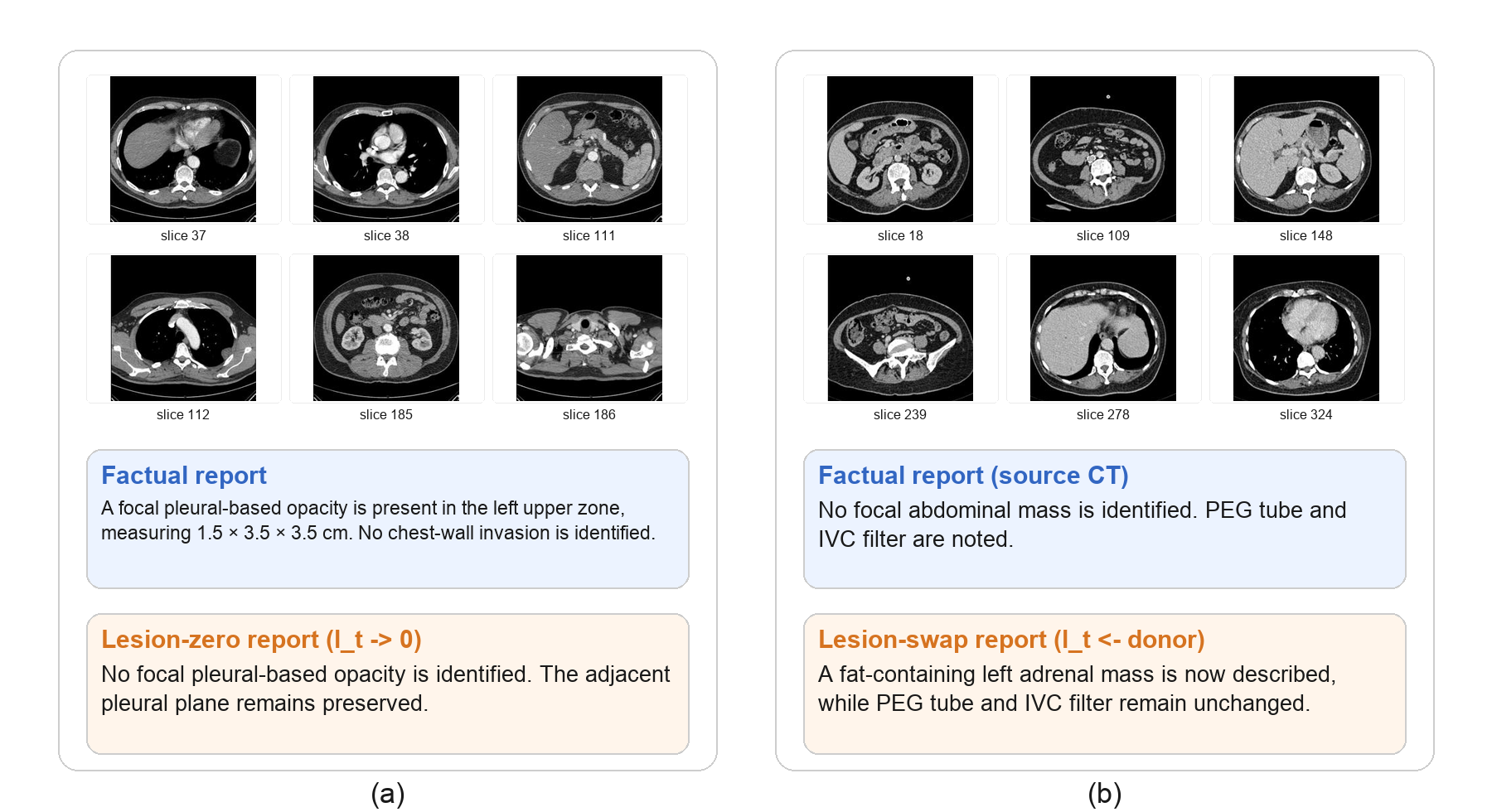}
\caption{Qualitative counterfactual lesion intervention examples. Panel (a) shows a lesion-positive case where lesion-zero removes a focal pleural-based opacity statement while preserving surrounding anatomical context. Panel (b) shows a lesion-factor insertion case where the inserted lesion factor introduces a donor-like adrenal-mass statement while keeping unrelated source findings unchanged.}
\label{fig:cf_examples}
\end{figure*}

\begin{figure*}[!h]
\centering
\includegraphics[width=\linewidth, trim=50 50 50 50]{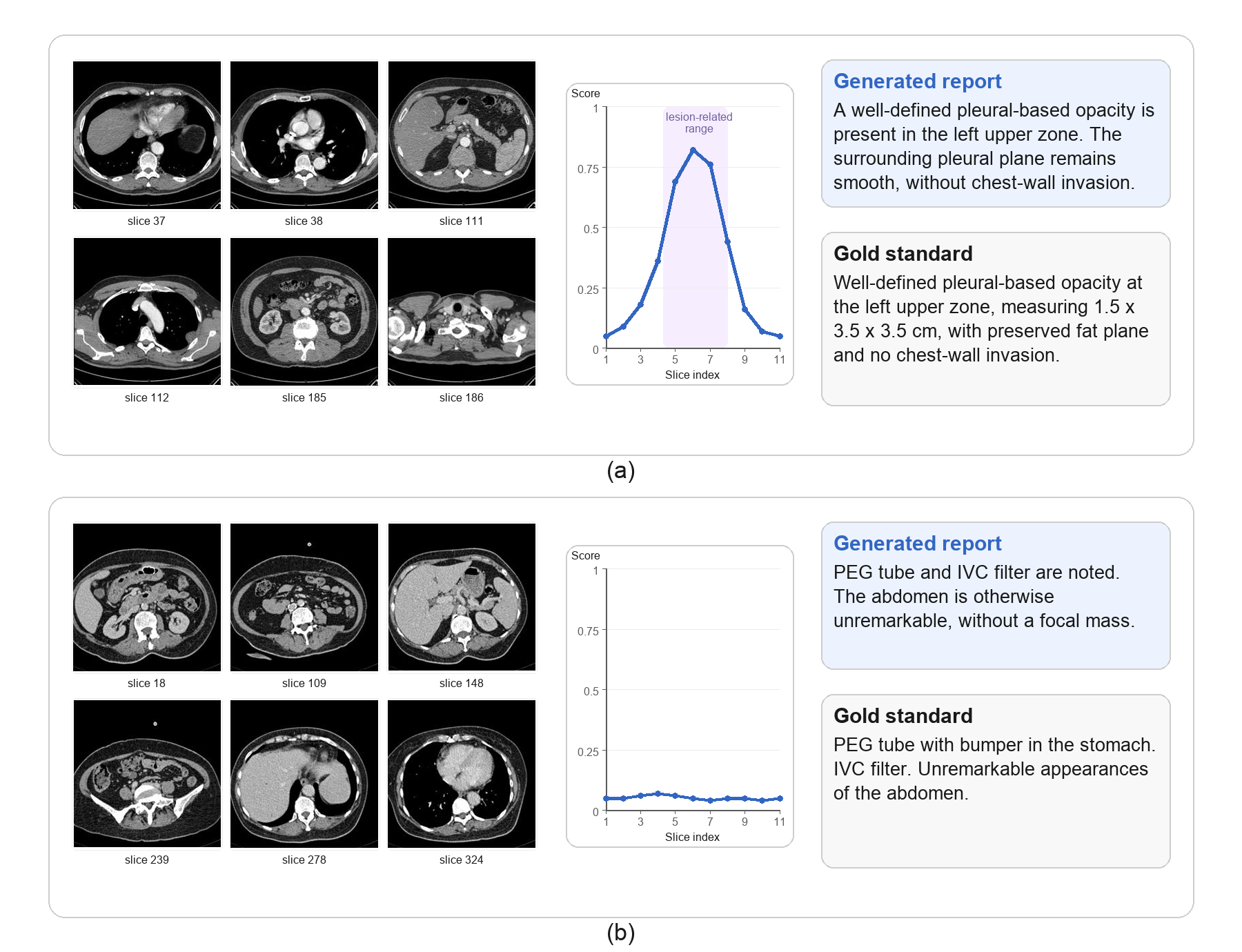}
\caption{Additional qualitative examples for factor-aware state behavior. The lesion-positive example shows a lesion trajectory rising over the slice interval associated with a focal pleural-based opacity, while the control example keeps the trajectory low and avoids hallucinating focal lesion terms.}
\label{fig:qualitative_appendix}
\end{figure*}

Figure \ref{fig:cf_examples} complements the selective lesion intervention analysis in Section 4.5 by showing representative lesion-zero and lesion-factor insertion examples.
Figure \ref{fig:qualitative_appendix} complements the factor-aware state analysis in Section 4.4 by visualizing representative CT slices, the lesion-factor trajectory, the generated report, and the reference report.
These qualitative figures are provided as illustrative examples; the corresponding aggregate claims are supported by the quantitative metrics in the main tables.

\section{Limitations}
\label{app:limitations}

SliceWorld uses the term world model in a restricted CT-specific sense: it learns a latent predictive state over ordered axial observations, predicts future slice-level visual evidence in feature space, and exposes factor-level interventions for report analysis.
It is not a HU-valued physical CT simulator, an action-conditioned RL environment model, or a model of disease progression over real time.
This operational definition is meaningful for CTRG because report generation depends on how local findings appear, persist, and disappear across the z-axis, and our study explores whether such axial evidence can be represented as a predictive and intervention-sensitive state rather than a direct volume-to-text feature.
Lesion-zero intervention should therefore be interpreted as latent lesion-factor neutralization rather than physical lesion removal from the image.
The factor-alignment analyses rely on available weak lesion labels, axial-position proxies, and automatic report labelers, so they support factor specialization rather than perfect disentanglement.
Our clinical evaluation combines retrospective public benchmarks, automatic metrics, and a 100-case evaluation by two clinical experts.
Multi-center external validation and prospective radiologist or workflow evaluation remain future directions.

\section{Ethics, Assets, and Release}
\label{app:ethics_assets}

This work uses publicly available, de-identified retrospective CT datasets and does not involve new data collection, crowdsourcing, or prospective human-subject experiments.
The method is intended as a research prototype for CT report generation and state-space analysis, not as a standalone clinical decision-making system.
Generated reports and counterfactual outputs should be reviewed by qualified clinicians before any practical use.
Potential risks include incorrect or incomplete generated findings, over-reliance on automated reports, and misinterpretation of latent lesion-factor interventions as physical image edits.
We mitigate these risks by evaluating clinically oriented metrics, reporting limitations, and framing latent lesion-factor report modulation as an analysis tool rather than a diagnostic intervention.

We use existing datasets and model components, including DeepLesion, M3D-Cap, CT-RATE, DINOv2, Mamba, Qwen3, and Ministral, and cite the corresponding original sources.
All assets are used under their original licenses, access terms, and dataset usage agreements.
We do not redistribute raw medical images, reports, or pretrained third-party model checkpoints.

\end{document}